\pdfoutput=1

\documentclass[11pt]{article}

\usepackage[preprint]{acl}

\usepackage{times}
\usepackage{latexsym}

\usepackage[T1]{fontenc}

\usepackage[utf8]{inputenc}

\usepackage{microtype}

\usepackage{inconsolata}

\usepackage{graphicx}
\usepackage{booktabs}
\usepackage{amsmath}
\usepackage{makecell}
\usepackage{colortbl}
\usepackage{array}
\usepackage{longtable}
\usepackage{amssymb}
\usepackage{subcaption}
\usepackage{multirow} 
\usepackage{listings}
\usepackage{xcolor}
\colorlet{numb}{magenta!60!black}
\lstdefinelanguage{json}{
    basicstyle=\ttfamily\small,
    numbers=left,
    numberstyle=\tiny\color{gray},
    stepnumber=1,
    numbersep=10pt,
    showstringspaces=false,
    breaklines=true,
    frame=single,
    literate=
     *{a}{{{\color{blue}a}}}{1}
      {b}{{{\color{blue}b}}}{1}
      {c}{{{\color{blue}c}}}{1}
      {d}{{{\color{blue}d}}}{1}
      {e}{{{\color{blue}e}}}{1}
      {f}{{{\color{blue}f}}}{1}
      {g}{{{\color{blue}g}}}{1}
      {h}{{{\color{blue}h}}}{1}
      {i}{{{\color{blue}i}}}{1}
      {j}{{{\color{blue}j}}}{1}
      {k}{{{\color{blue}k}}}{1}
      {l}{{{\color{blue}l}}}{1}
      {m}{{{\color{blue}m}}}{1}
      {n}{{{\color{blue}n}}}{1}
      {o}{{{\color{blue}o}}}{1}
      {p}{{{\color{blue}p}}}{1}
      {q}{{{\color{blue}q}}}{1}
      {r}{{{\color{blue}r}}}{1}
      {s}{{{\color{blue}s}}}{1}
      {t}{{{\color{blue}t}}}{1}
      {u}{{{\color{blue}u}}}{1}
      {v}{{{\color{blue}v}}}{1}
      {w}{{{\color{blue}w}}}{1}
      {x}{{{\color{blue}x}}}{1}
      {y}{{{\color{blue}y}}}{1}
      {z}{{{\color{blue}z}}}{1}
      {.}{{{\color{black}.}}}{1}
      {:}{{{\color{blue}:}}}{1}
      {\_}{{{\color{blue}\_}}}{1}
      {"}{{{\color{black}"}}}{1}
      {,}{{{\color{black},}}}{1}
      {\{}{{{\color{black}\{}}}{1}
      {\}}{{{\color{black}\}}}}{1}
      {[}{{{\color{black}[}}}{1}
      {]}{{{\color{black}]}}}{1},
}

\title{Enriching Patent Claim Generation with European Patent Dataset}

\author{Lekang Jiang, Chengzu Li, Stephan Goetz \\
        University of Cambridge \\
        \texttt{\{lj408, cl917, smg84\}@cam.ac.uk}}

\begin{document}
\maketitle

\begin{abstract}

Drafting patent claims is time-intensive, costly, and requires professional skill. Therefore, researchers have investigated large language models (LLMs) to assist inventors in writing claims. However, existing work has largely relied on datasets from the United States Patent and Trademark Office (USPTO). To enlarge research scope regarding various jurisdictions, drafting conventions, and legal standards, we introduce EPD, a European patent dataset. EPD presents rich textual data and structured metadata to support multiple patent-related tasks, including claim generation.
This dataset enriches the field in three critical aspects: 
\textbf{(1) Jurisdictional diversity:} Patents from different offices vary in legal and drafting conventions. EPD fills a critical gap by providing a benchmark of European patents to enable more comprehensive evaluation.
\textbf{(2) Quality improvement:} EPD offers high-quality granted patents with finalized and legally approved texts, whereas others consist of patent applications that are unexamined or provisional. Experiments show that LLMs fine-tuned on EPD significantly outperform those trained on previous datasets and even GPT-4o in claim quality and cross-domain generalization.
\textbf{(3) Real-world simulation:} We propose a difficult subset of EPD to better reflect real-world challenges. Results reveal that all tested LLMs perform substantially worse on challenging samples, which highlights the need for future research. \footnote{\url{https://github.com/scylj1/EPD}}

\end{abstract}

\begin{table*}[!ht]
\centering
\footnotesize
\resizebox{\textwidth}{!}{
\begin{tabular}{l|c|cccccc|c|c|c|c}
\toprule
    \textbf{Dataset} & \textbf{Size} & \textbf{Title} & \textbf{Abst} & \textbf{Claim}& \textbf{Des} & \textbf{Cit}  & \textbf{PC}  & \textbf{Years}  & \textbf{Source} & \textbf{Version} & \textbf{Primary Purpose} \\
\midrule
WIPO-alpha & 75K & \checkmark & \checkmark & \checkmark & \checkmark & & \checkmark & 1998-2002 & WIPO & Appl & Classification \\
USPTO-2M & 2M & \checkmark & \checkmark & \checkmark & & & \checkmark & 2006-2015 & USPTO & Gra & Classification \\
PatentMatch & 6.3M &  &  & \checkmark & & \checkmark &  & <2020 & EPO & Appl \& Gra & Retrieval \\
BigPatent & 1.3M & \checkmark & \checkmark & & \checkmark & &  & 1971-2018 & Google & Gra & Summarization \\
HUPD-DCG & 9.5K & \checkmark & \checkmark & \checkmark & \checkmark & & \checkmark & 2017 & USPTO & Appl & Claim generation \\
Patent-CR & 22.6K &  &  & \checkmark &  & &  & 2024 & EPO & Appl \& Gra & Claim revision \\
HUPD & 4.5M & \checkmark & \checkmark & \checkmark & \checkmark & & \checkmark & 2004-2018 & USPTO & Appl & Multi-purpose \\
\rowcolor{gray!30}
EPD & 107K & \checkmark & \checkmark & \checkmark & \checkmark & \checkmark& \checkmark & 2024 & EPO & Appl \& Gra & Multi-purpose \\
\bottomrule
\end{tabular}}
\caption{Comparison of related patent datasets. Abbreviations: Abst -- Abstract, Dsc -- Description, Cit --- Citation, PC -- IPC/CPC codes, Appl -- Application, and Gra -- Granted version.}
\label{table:datasets}
\end{table*}

\section{Introduction}

Patents grant inventors temporary exclusive rights to prevent competitors from using the invention in return for innovation and its detailed public disclosure \citep{Frumkin1947}. Patent claims define the legal boundaries of an invention, which renders them one of the most critical components of patent documents \citep{epo2020}. High-quality claims must balance comprehensiveness and enforceability to ensure robust protection while adhering to legal and regulatory standards. The complexity of drafting and revising patent claims makes the process both time-consuming and costly \citep{jiang-etal-2025-large}. We include more detailed background information on patents in Appendix \ref{background}. 

Recent advances in natural language processing (NLP) and large language models (LLMs) have shown promise in general text generation \citep{zhao2023survey,li-etal-2025-large-language-models} and patent-related tasks \citep{jiang2025natural}. Researchers have explored LLMs to assist inventors and patent professionals in automatically generating well-structured claims. Early approaches primarily focused on fine-tuning neural models to generate claims from abstracts \citep{lee2020patentgenerate, lee2020controlling}. However, \citet{jiang-etal-2025-large} found that abstract-based claim generation often leads to suboptimal results due to the generic nature of abstracts. Consequently, recent work and the increase of context length in LLMs have enabled description-based claim generation, which can leverage the detailed technical content in patent descriptions to enhance claim completeness and quality \citep{jiang-etal-2025-large, wang2024autopatent,jiang-etal-2025-towards-better}.

Although prior studies have made significant progress, they are only based on patents from the United States Patent and Trademark Office (USPTO). However, patents from different jurisdictions can vary significantly in linguistic style and legal drafting conventions because of distinct examination practices and legal frameworks. For example, USPTO claims emphasize broad coverage, while European claims tend to be more concise and precise, with stricter requirements on clarity and support in the description. These differences not only impact how claims are written and interpreted but also pose challenges for training language models. Thus, datasets from different patent offices are essential to comprehensively evaluate LLMs' performance and robustness across diverse legal and linguistic contexts.

This paper introduces EPD, a dataset of English-language patents granted by the European Patent Office (EPO) in 2024. Compared to existing datasets for patent claim generation, EPD offers three major advantages: 

\noindent \textbf{Higher Quality:} EPD contains granted patents directly sourced from the EPO to ensure high-quality and legally validated claims. In contrast, as shown in Table \ref{table:datasets}, prior datasets that support claim generation only include the (accepted, not redacted) application versions, which may compromise data quality. For example, they may still include canceled claims and be different from granted versions. 

\noindent \textbf{Closer to Real-World Scenarios:} EPD includes a difficult subset designed to simulate real-world situations of claim drafting. This subset enables a more rigorous evaluation of LLMs’ ability to generate high-quality claims under practical conditions. 

\noindent \textbf{Lower Risk of Data Leakage:} EPD consists of patents granted in 2024, while existing datasets that support claim generation are before 2018 (see Table~\ref{table:datasets}). This temporal gap reduces the likelihood that the data overlaps with the pre-training corpora of current LLMs.

Our main contributions are detailed as follows: 

\noindent 1. We present the EPD dataset, a collection of 107K European patents with rich textual data and structured metadata to support various patent-related tasks. EPD addresses a key gap in the field by enabling cross-jurisdiction evaluation of LLMs. While our primary focus is on claim generation, we also highlight other potential applications in Appendix~\ref{moretask}.

\noindent 2. We conduct a comprehensive evaluation of several LLMs on both the HUPD-DCG dataset (from USPTO) \citep{jiang-etal-2025-large} and our EPD dataset (from EPO). Results show that LLMs fine-tuned on EPD significantly outperform those trained on HUPD-DCG and even GPT-4o across both test sets. It indicates EPD’s effectiveness in claim generation and cross-domain generalization.

\noindent 3. We introduce a difficult subset of EPD to mimic real-world challenges in claim generation. Experimental results reveal a marked performance drop across all tested LLMs on difficult samples, which highlights the need for future research.

\section{Related Work}

\subsection{Patent Claim Generation}

Claim generation is a crucial aspect of patent drafting as it defines the boundaries of the patent's protection. The task of claim generation involves formulating precise, comprehensive claims that capture the essence of the invention while ensuring enforceability. Early studies fine-tuned models to generate claims based on patent abstracts \citep{lee2020patentgenerate, lee2020controlling}. However, \citet{jiang-etal-2025-large} revealed that abstract-based claim generation may not be a well-conditioned task, because abstracts are usually generic and imprecise. Hence, they proposed the description-based claim generation task and evaluated the performance of different LLMs with patents from USPTO. This paper conducted a more comprehensive and deeper analysis for patent claim generation, which offers valuable insights for future research.

\subsection{Patent Datasets}

Curated patent datasets are important for researchers to develop LLMs and NLP tools to explore this specific field. As shown in Table~\ref{table:datasets}, previous studies have created some patent datasets, including WIPO-alpha \citep{fall2003automated} and USPTO-2M \citep{li2018deeppatent} for patent classification, PatentMatch \citep{risch2020patentmatch} for retrieval, BigPatent \citep {sharma-etal-2019-bigpatent} for summarization, HUPD-DCG \citep{jiang-etal-2025-large} for claim generation, and Patent-CR \citep{jiang-etal-2025-patent} for claim revision. In addition, HUPD \citep{suzgun2024harvard} is a recent large-scale dataset supporting multiple tasks, such as the classification of patent decisions and patent summarization. Our EPD is the first large-scale multi-purpose dataset based on European patents.

\section{Dataset}

\subsection{Construction}

The EPO offers Open Patent Services (OPS), which provides public access to their data.\footnote{\url{https://www.epo.org/en/searching-for-patents/data/web-services/ops}} The EPO publishes different versions of a patent, where the \textit{A} code represents the published patent application, and the \textit{B} code indicates the granted patent.\footnote{\url{https://se.espacenet.com/help?locale=en_SE&method=handleHelpTopic&topic=kindcodes}}
We retrieve English-based European patents granted between January 2024 and August 2024 by EPO. 
Both the application and granted versions of patents are retrieved through the OPS API. We kept the granted version if no related applications were provided. 
The retrieved data was in XML format. We removed patent files if some of the required fields were missing. We processed and organized the data into a structured and easily readable JSON format. An example data file is shown in Figure \ref{fig:exampledata}.

\subsection{Statistics}

Table~\ref{table:dataset_statistics} presents the dataset statistics of EPD. It includes 73,980 patents in total, of which 32,988 have both the application and granted versions. The granted versions contain fewer tokens than the application version in corresponding sections on average, suggesting that the granted versions are usually more concise. Notably, the patent descriptions often exceed 10,000 tokens, which poses a challenge for the context length limitations of some LLMs. 

\begin{table}[t!]
\centering
\footnotesize

\begin{tabular}{l|c|c}
\toprule
    & \textbf{Application} & \textbf{Granted} \\ \midrule
\textbf{\# Patents} & \multicolumn{2}{c}{} \\
\# Patents (all) & \multicolumn{2}{c}{73,980}   \\
\# Patents (incl. both version) & \multicolumn{2}{c}{32,988} \\ \midrule
\textbf{\# Documents} & 32,988 & 73,980  \\ \midrule
\textbf{Average \# Tokens} & \multicolumn{2}{c}{} \\
Title  &  15.8  & 15.5    \\
Abstract  &  165.1  &  --    \\
Claim  & 1372.1 &  1271.2   \\
Description & 15280.5 & 14320.5 \\ \bottomrule
\end{tabular}

\caption{Dataset statistics. The granted patents from EPO in this dataset do not include the abstract section.  }
\label{table:dataset_statistics}
\end{table}

\section{Experiments}

\subsection{Datasets for Claim Generation}

\textbf{Fine-Tuning and Test Datasets} 
This study mainly focuses on the patent claim generation task. We use two datasets for comparison, \textbf{HUPD-DCG} \citep{jiang-etal-2025-large} and our proposed \textbf{EPD} dataset. 
HUPD-DCG consists of patent documents filed in 2017 and granted by the USPTO. To accommodate the context length of some LLMs and reduce computational costs, HUPD-DCG only includes patents with description lengths under 8,000 tokens. To ensure a fair comparison, we apply the same filtering criterion to EPD and obtain 8,007 samples, which contain only granted patents from EPO. We allocate 1,035 samples (approximately 13\%) for testing, while the remaining 6,972 samples form the training set. To maintain dataset consistency, we randomly select examples from the original HUPD-DCG dataset. Both training sets (HUPD-DCG and EPD) contain 6,972 examples, and both test sets include 1,035 examples.

Additionally, to explore the effectiveness of mixed fine-tuning, we construct a \textbf{mixed dataset} that comprises 3,486 randomly selected training samples from HUPD-DCG and 3,486 from EPD.

\noindent \textbf{Comparison between EPD and HUPD-DCG}
Patents from different jurisdictions vary in their drafting convention and legal standards. To analyze the detailed differences, we calculate the claim statistics of two datasets in Table \ref{table:rq1_statistics}, including basis statistics, structural \& syntactic complexity, and linguistic \& stylistic features. 
While the average number of claims per patent is slightly lower in EPD (11.5 vs.\ 13.6), the average length per claim is higher (84.5 vs.\ 81.1 tokens). This indicates that EPO claims tend to be more elaborate, with a greater focus on comprehensive coverage within fewer claims. EPD also contains fewer independent and dependent claims on average, which reflects jurisdictional differences in claim structuring. 
Structurally, EPD exhibits deeper syntactic trees (47.0 vs. 38.2), but lower structural complexity (2.35 vs. 2.77) and better readability (26.8 vs. 30.1), which suggests that EPO claims are more syntactically nested yet structurally clearer and more readable. 
Linguistically, EPD shows higher lexical diversity (30.6 vs. 28.2) and term density (0.20 vs. 0.10), which reflects denser and more varied technical language. 
The variations in structure and linguistic patterns can influence how models process and generate claims, which can lead to different performance of the same model on two datasets.

\noindent \textbf{Difficult Subset Creation}
\citet{jiang-etal-2025-large} found that LLMs can generate high-quality first independent claims primarily because these claims often appear verbatim in the description so that models can identify and extract rather than generate them. In real-world scenarios, patent attorneys usually need to identify key technical features of the invention---which should be described in detail in the description---and reconstruct them into well-structured claims. Therefore, to assess LLMs' true claim generation capabilities beyond simple extraction, we classify samples into easy and difficult categories and evaluate them separately.

A patent is classified as easy if at least one sentence in the description has an ROUGE-L \citep{lin2004rouge} score above 0.6 with the first claim. We set this threshold because LLMs generally struggle to achieve an ROUGE-L above 0.6, as reported in previous studies \cite{jiang-etal-2025-large} and shown in Table \ref{table:rq1_main}. If no sentence meets this criterion, the patent is classified as difficult because there is no significant overlap for models to do simple extraction. Based on this classification, we identify 693 difficult and 342 easy samples in EPD's test set.

\subsection{Models}
We select \textbf{Llama-3.1-8B}\footnote{\url{https://huggingface.co/meta-llama/Llama-3.1-8B-Instruct}} \citep{dubey2024llama} as the base model because of its proven capabilities and publicly availability for fine-tuning. We also explore the few-shot alternatives, where the model is given a single illustrative example. In addition, we test with \textbf{GPT-4o}\footnote{GPT-4o-2024-08-06: \url{https://platform.openai.com/docs/models/gpt-4o}} and \textbf{DeepSeek-V3} \citep{deepseekai2024deepseekv3technicalreport} as strong baselines. We evaluate their zero-shot performance on the claim generation task. 
We do not test patent-related LLMs because current models are either not publicly available \citep{bai2024patentgpt} or have demonstrated suboptimal performance on claim generation tasks \citep{lee2023evaluating, jiang-etal-2025-large}. In addition, we evaluate the legal-domain LLM \textbf{SaulLM-7B}\footnote{\url{https://huggingface.co/Equall/Saul-7B-Instruct-v1}} \citep{colombo2024saullm} for comparison with prior studies \citep{jiang-etal-2025-large}.

We further fine-tune the Llama-3.1-8B model with LoRA \citep{hu2021lora} on the HUPD-DCG, EPD, and mixed datasets, which produce three fine-tuned models: \textbf{Llama-3.1-8B-FT (HUPD-DCG)}, \textbf{Llama-3.1-8B-FT (EPD)}, and \textbf{Llama-3.1-8B-FT (Mixed)}. Appendix \ref{expdetails} reports experimental details. 

\begin{table}[t!]
\centering
\footnotesize
\begin{tabular}{l|p{1.8cm}<{\centering}|p{1.8cm}<{\centering}}
\toprule
\textbf{Statistics} & \textbf{HUPD-DCG} & \textbf{EPD} \\
\midrule

\multicolumn{3}{l}{\textbf{Basic Statistics}} \\
\# Tokens &  926 $\pm$ 495 & 929 $\pm$ 381 \\
\# Claims  & 13.6 $\pm$ 6.9  & 11.5 $\pm$ 3.9 \\
\# Independent Claims & 1.9 $\pm$ 0.9 & 1.4 $\pm$ 0.6 \\
\# Dependent Claims & 11.7 $\pm$ 6.5 & 10.1 $\pm$ 3.8 \\
Per Claim Length  & 81.1 $\pm$ 52.5 & 84.5 $\pm$ 31.6 \\ 
\midrule
\multicolumn{3}{l}{\textbf{Structural \& Syntactic Complexity}} \\
Structure Complexity  & 2.77 $\pm$ 2.69 & 2.35 $\pm$ 1.44 \\
Syntactic Tree Depth & 38.2 $\pm$ 19.7 & 47.0 $\pm$ 20.8 \\
Readability ($\downarrow$)   & 30.1 $\pm$ 18.8 & 26.8 $\pm$ 9.4 \\
\midrule
\multicolumn{3}{l}{\textbf{Linguistic \& Stylistic Features}} \\
Lexical Diversity  & 28.2 $\pm$ 5.5 & 30.6 $\pm$ 6.2 \\
Term Density & 0.10 $\pm$ 0.06 & 0.20 $\pm$ 0.08 \\

\bottomrule
\end{tabular}
\caption{Comparison of claim statistics with standard deviations between HUPD-DCG and EPD. The methods to calculate these statistics are introduced in Appendix~\ref{statscalculation}. A smaller readability score indicates higher readability. }
\label{table:rq1_statistics}
\end{table}

\begin{table*}[ht!]
\footnotesize
\centering
\resizebox{\linewidth}{!}{
\begin{tabular}{lcccccccccc}
\toprule
& \multicolumn{4}{c}{\textbf{EPD}} & \multicolumn{4}{c}{\textbf{HUPD-DCG}}  \\
\cmidrule(lr){2-5} \cmidrule(lr){6-9}
\textbf{Model} & \textbf{BLEU} & \textbf{R-1} & \textbf{R-L} & \textbf{BERTScore} & \textbf{BLEU} & \textbf{R-1} & \textbf{R-L} & \textbf{BERTScore} \\
\midrule

\multicolumn{3}{l}{\textbf{Law-specific LLMs}} \\
SaulLM-7B & 13.53 & 39.95 & 26.46 & 83.19 & 12.68* & 36.63* & 25.10* & 83.13*  \\
\midrule

\multicolumn{3}{l}{\textbf{Base LLMs}} \\
Llama-3.1-8B (Zero-shot) & 28.52 & 60.76 & 39.58 & 86.32 & 34.30 & 59.79 & 40.91 & 87.64 \\
Llama-3.1-8B (Few-shot of HUPD)	& 25.69	& 65.16 & 41.73 & 86.43 & 28.61 & \textbf{63.46} & 41.73 & 86.43 \\
Llama-3.1-8B (Few-shot of EPD)	& 33.12	& 62.75 & 43.92 & 87.78 & 34.96 & 62.16 & 43.92 & 87.78 \\
GPT-4o (Zero-shot) & 21.26 & 60.21 & 40.17 & 85.95 & 26.57 & 58.21 & 40.21 & 87.25 \\
DeepSeek-V3 (Zero-shot)	& 36.53 & 65.70 & 45.60 & 86.43 & 39.66 & 62.63 & 44.66 & 87.06 \\
\midrule

\multicolumn{3}{l}{\textbf{Fine-tuned LLMs}} \\
Llama-3.1-8B-FT (HUPD-DCG) & 36.02 & 63.02 & 47.76 & 88.30 & 40.54 & 61.89 & \textbf{48.70} & 89.89 \\
Llama-3.1-8B-FT (EPD) & \textbf{47.22} & \textbf{67.94} & \textbf{51.52} & \textbf{90.40} & 34.05 & 58.95 & 45.16 & 87.99  \\
Llama-3.1-8B-FT (Mixed) & 47.04  & 67.11 & 50.87 & 90.31 & \textbf{41.50} & 61.61 & 47.94 & \textbf{89.92}  \\

\bottomrule
\end{tabular}
}
\caption{Claim generation results on EPD (from EPO) and HUPD-DCG (from USPTO) datasets. The best scores for each metric are marked in \textbf{bold}. Scores of * are from \citet{jiang-etal-2025-large}. }
\label{table:rq1_main}
\end{table*}

\subsection{Evaluation Metrics}
We adopt traditional standard evaluation metrics for text generation, including BLEU \citep{papineni2002bleu}, ROUGE-1 (R-1), ROUGE-L (R-L) \citep{lin2004rouge}, and BERTScore \citep{zhang2019bertscore}. BLEU, R-1, and R-L assess surface-level text similarity by measuring word or sequence overlap between generated outputs and reference texts, while BERTScore captures semantic similarity.

Moreover, studies have shown that LLM-as-a-judge evaluators can achieve better human alignment \citep{liu-etal-2023-g} and also in patent claim evaluation \citep{jiang-etal-2025-patent}. Thus, we use Deepseek-V3 \citep{liu2024deepseek} with Chain-of-Thought (CoT) \citep{wei2022chain} prompting to evaluate generated patent claims. We do not evaluate the generation of itself to ensure fairness.
The evaluation dimensions include: feature coverage, technical precision, specificity, clarity, dependency structure, legal terminology, and scope of protection.  
Detailed settings and prompts are introduced in Appendix~\ref{llmasjudge}. 

\section{Results}

\begin{figure*}[!t]
    \centering   
    \includegraphics[width=\textwidth]{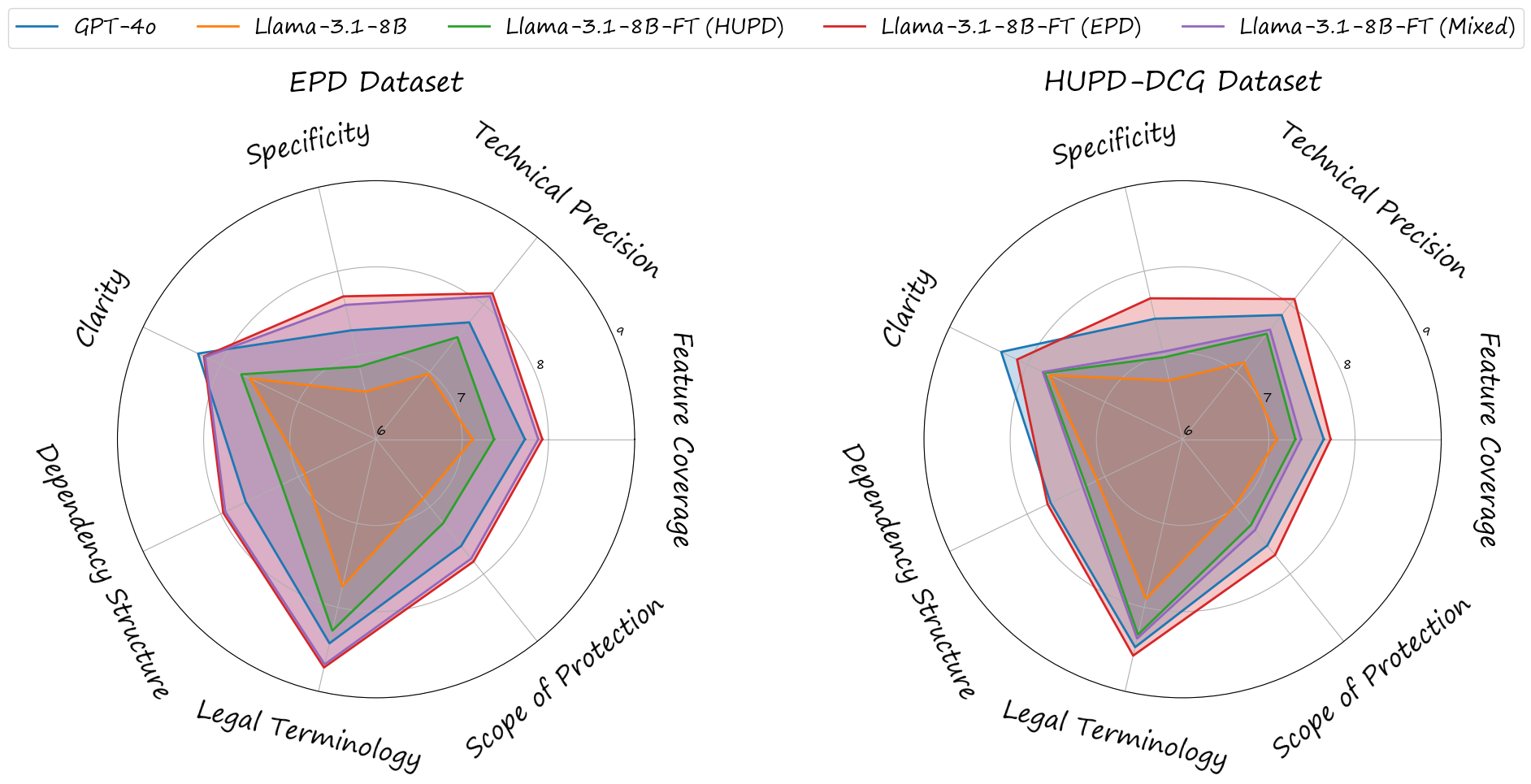}   
    \caption{LLM-as-a-judge evaluation results on EPD and HUPD-DCG.}
    \label{fig:epo+us}
\end{figure*}

\subsection{Performance on European and US Patent Datasets}

Table \ref{table:rq1_main} and Figure~\ref{fig:epo+us} respectively present the traditional and LLM-as-a-judge evaluation results of different LLMs on two datasets, EPD (patents from EPO) and HUPD-DCG (patents from USPTO). 

Previous research has shown that fine-tuning LLMs on a specific dataset significantly improves claim generation quality, including the completeness of invention features, conceptual clarity, and feature linkage \citep{jiang-etal-2025-large}. Therefore, our study primarily focuses on model performance across different datasets,  model generalization ability, and linguistic differences between datasets. Additionally, \citet{jiang-etal-2025-large} reported the poor performance of legal-specific LLM SaulLM-7B on claim generation with detailed explanations. Our results in Table \ref{table:rq1_main} further confirm its ineffectiveness on the EPD dataset, where it significantly lags behind other models. For example, SaulLM-7B achieves an R-L score of 26.46 on EPD, more than 13\% lower than Llama-3.1-8B. Since our experiments on EPD revealed similar findings, we exclude further discussions of SaulLM-7B.

\noindent \textbf{The EPD-trained model significantly improves claim generation quality.}
Table \ref{table:rq1_main} shows that fine-tuning on the EPD dataset leads to substantial improvements across both traditional metrics (BLEU, ROUGE, BERTScore) and LLM-as-a-judge evaluations (e.g., clarity, dependency, terminology) on the EPD test set. Compared to other models, Llama-3.1-8B-FT (EPD) achieves the highest BLEU (47.22), R-1 (67.94), R-L (51.52), and BERTScore (90.40). Similarly, as shown in Figure~\ref{fig:epo+us}, Llama-3.1-8B-FT (EPD) demonstrates stronger LLM-based scores across all dimensions, with the highest scores in content (7.9), precision (8.2), specificity (7.7), dependency (8.0), terminology (8.7), and scope (7.8). Notably, GPT-4o achieves the highest clarity score of 8.3. In addition, while fine-tuning on HUPD-DCG also improves over the base model, the gains are more modest. 
Under few-shot settings, we observe similar patterns to those seen in the fine-tuning setup: one-shot prompts drawn from EPD lead to better performance than those from HUPD, on both the EPD and HUPD test sets. 

\begin{figure}[!t]
    \centering   
    \includegraphics[width=.49\textwidth]{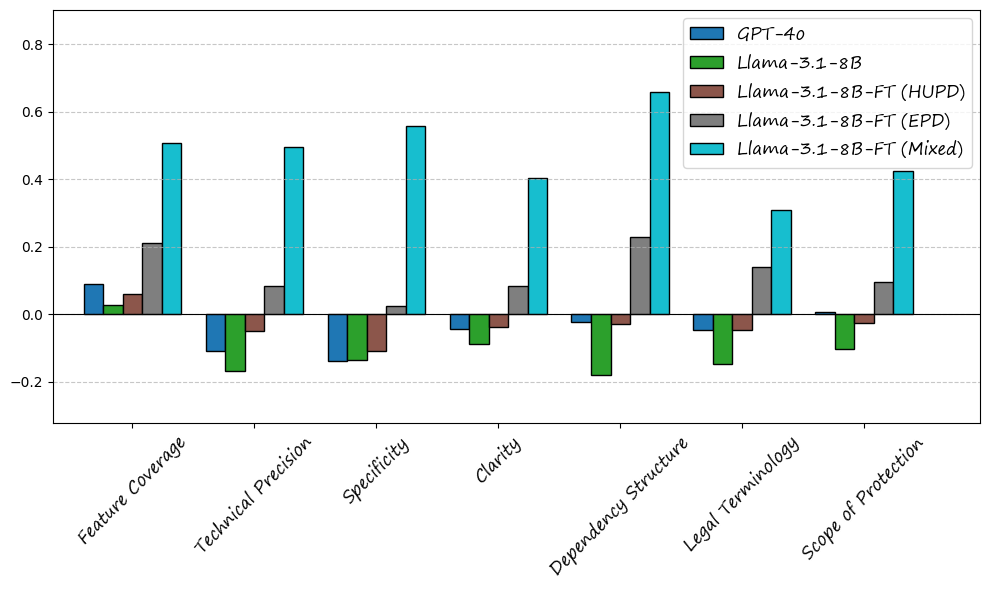} 
    \caption{LLM-as-a-judge performance difference between EPD and HUPD-DCG (EPD minus HUPD-DCG)}
    \label{fig:epo-us}
\end{figure}

\noindent \textbf{Base LLMs perform worse on EPD.}
Figure~\ref{fig:epo-us} highlights the performance gap between EPD and HUPD-DCG. Base models (Llama-3.1-8B and GPT-4o) generally perform better on HUPD-DCG, with performance dropping by up to 0.2 points on most evaluation dimensions, except for content coverage. This may be due to: (1) extensive pre-training on US patent data, leading to a preference for US-style claims; and (2) the greater precision and structural rigor of granted claims in EPD, which pose challenges for zero-shot generation. Overall, these findings underscore the distinctiveness of EPD and its value in advancing claim generation research.

\begin{figure*}[!t]
    \centering   
    \includegraphics[width=\textwidth]{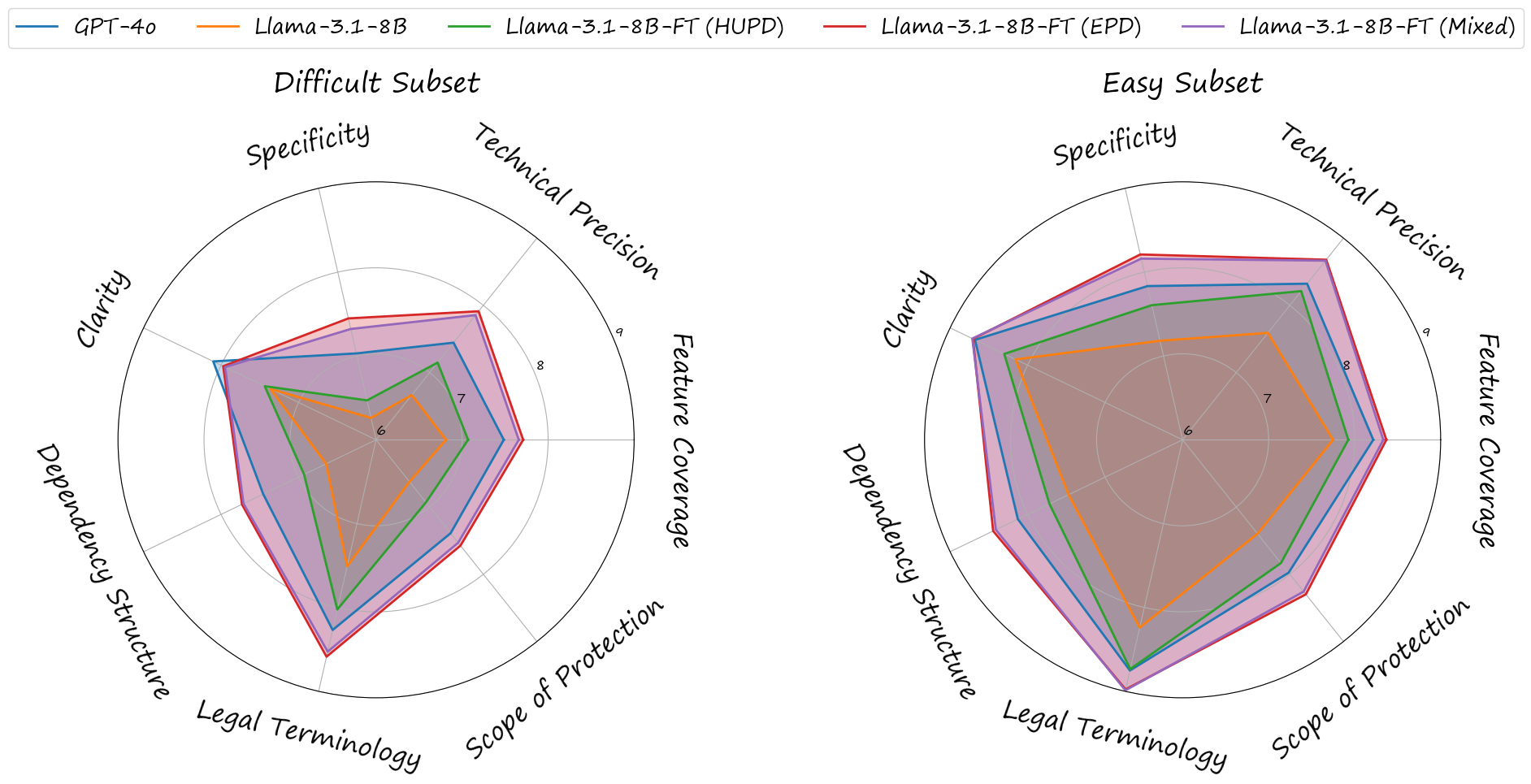}   
    \caption{LLM-as-a-judge performance on difficult and easy subsets of EPD.}
    \label{fig:easy+diff}
\end{figure*}

\noindent \textbf{The EPD-trained model exhibits cross-domain generalization ability.}
As shown in Figure~\ref{fig:epo-us}, models fine-tuned on one dataset experience a drop when tested on the other dataset. For example, the performance of Llama-3.1-8B-FT (EPD) degrades in all aspects when evaluated on HUPD-DCG, especially in content coverage and dependency (more than 0.2). The performance drops across datasets suggest the generalization challenge, as patents from the USPTO and EPO have different drafting conventions and standards. 
However, Llama-3.1-8B-FT (EPD) still exceeds Llama-3.1-8B-FT (HUPD-DCG) on LLM-based evaluation on the HUPD-DCG dataset, as shown in Figure \ref{fig:easy+diff}. Llama-3.1-8B-FT (EPD) achieves higher scores in content (7.7 vs. 7.3), precision (8.1 vs. 7.6), specificity (7.7 vs. 7.0), clarity (8.1 vs 7.8), dependency (7.7 vs. 7.2), terminology (8.6 vs. 8.3), and scope (7.7 vs. 7.3). This result implies that higher-quality training data can potentially enable stronger generalization across patent jurisdictions.

On the other hand, in Table \ref{table:rq1_main}, Llama-3.1-8B-FT (HUPD-DCG) demonstrates slightly higher traditional evaluation scores on the HUPD-DCG test set. Specifically, Llama-3.1-8B-FT (HUPD-DCG) achieves higher R-1 (61.89 vs. 58.95), R-L (48.70 vs. 45.16), and BERTScore (89.89 vs. 87.99). This is possibly because the fine-tuned model mimics the verbose US-style ground truth more closely. Nonetheless, LLM-based judgments highlight that Llama-3.1-8B-FT (EPD) generates claims that are more faithful to the legal and structural norms of gold claims.

\noindent \textbf{Mixed-domain training does not bring extra benefits.}
The model fine-tuned on the mixed dataset (EPD + HUPD-DCG) achieves comparable performance to Llama-3.1-8B-FT (EPD) on the EPD test set and to Llama-3.1-8B-FT (HUPD-DCG) on the HUPD-DCG test set, across both traditional metrics and LLM-as-a-judge evaluations. For instance, BLEU, ROUGE, and BERTScore variations are within 1\% compared to the models fine-tuned on individual datasets. As shown in Figure~\ref{fig:epo+us}, the performance profiles of Llama-3.1-8B-FT (Mixed) nearly overlap with those of the corresponding fine-tuned models on their respective test sets.
However, its LLM-as-a-judge scores remain slightly lower than those of Llama-3.1-8B-FT (EPD) on both datasets, with a marginal drop (less than 0.1) on EPD, but a more noticeable decrease (up to 0.5) on HUPD-DCG. Moreover, Figure~\ref{fig:epo-us} reveals that mixed training results in larger cross-domain performance drops, exceeding 0.3 in all evaluation dimensions. These findings suggest that mixed training does not help the model acquire additional structural or linguistic knowledge to improve claim generation quality or cross-domain generalization.

\begin{table*}[ht!]
\footnotesize
\centering
\resizebox{\linewidth}{!}{
\begin{tabular}{lcccccccccc}
\toprule
& \multicolumn{4}{c}{\textbf{Difficult Samples}} & \multicolumn{4}{c}{\textbf{Easy Samples}}  \\
\cmidrule(lr){2-5} \cmidrule(lr){6-9}
\textbf{Model} & \textbf{BLEU} & \textbf{R-1} & \textbf{R-L} & \textbf{BERTScore} & \textbf{BLEU} & \textbf{R-1} & \textbf{R-L} & \textbf{BERTScore} \\
\midrule

\multicolumn{3}{l}{\textbf{Base LLMs}} \\
Llama-3.1-8B (Zero-shot) & 24.41 & 58.54 & 35.90 & 85.46 & 37.14 & 65.21 & 47.04 & 88.04 \\
Llama-3.1-8B (one-shot of HUPD) & 21.77 & 56.75 & 37.31 & 85.54 & 34.18 & 64.13 & 50.77 &	88.22\\
Llama-3.1-8B (one-shot of EPD)	& 28.85 & 60.24  & 38.98 & 86.77 & 42.24 & 67.85 & 53.96 &	89.81  \\
GPT-4o (Zero-shot) & 17.99 & 57.72 & 36.42 & 85.20 & 28.43 & 65.19 & 47.72 & 87.48 \\
DeepSeek-V3 (Zero-shot) & 32.14 & 63.35 & 40.90 & 85.55 & 44.79 & 70.43 & 55.01 & 88.21	\\ 
\midrule

\multicolumn{3}{l}{\textbf{Fine-tuned LLMs}} \\
Llama-3.1-8B-FT (HUPD-DCG) & 31.45 & 59.61 & 42.29 & 87.13 & 45.77 & 69.91 & 58.90 & 90.69 \\
Llama-3.1-8B-FT (EPD) & 42.62 & \textbf{64.93} & \textbf{45.83} & \textbf{89.09} & \textbf{57.01} & \textbf{74.10} & \textbf{63.08} & \textbf{93.05} \\
Llama-3.1-8B-FT (Mixed) & \textbf{42.66} & 64.26 & 45.68 & 89.04 & 56.45& 72.97 & 61.50 & 92.86 \\

\bottomrule
\end{tabular}
}
\caption{Claim generation results on difficult and easy examples of the EPD dataset. The best scores for each metric are marked in \textbf{bold}. }
\label{table:rq2_main2}
\end{table*}

\noindent \textbf{High-quality training data are essential for improving both generation quality and generalization.}
Although HUPD-DCG originates from USPTO, it consists of application-stage claims, which are generally longer and structurally more complex, as shown in Table~\ref{table:rq1_statistics}.  As demonstrated in Table~\ref{table:rq1_main} and Figure~\ref{fig:epo+us}, fine-tuning on such data leads to limited performance improvements and fails to generalize to other datasets.
In contrast, EPD contains granted claims from EPO, which are more compact, terminology-rich, and legally polished. The polished granted claims allow models to learn more stable linguistic and logical patterns, which results in better claim generation performance in both in-domain and cross-domain settings, even surpassing GPT-4o. These findings highlight the critical role of high-quality data in enabling robust and precise legal text generation.

\subsection{Performance on Difficult and Easy Samples of EPD}

Table \ref{table:rq2_main2} and Figure \ref{fig:easy+diff} respectively present the traditional and LLM-as-a-judge results for difficult and easy subsets of EPD.

\noindent \textbf{The EPD-trained model consistently outperforms others on both easy and difficult subsets.}
Consistent with earlier findings, Llama-3.1-8B fine-tuned on EPD achieves the best results across both subsets and nearly all metrics. As shown in Table \ref{table:rq2_main2}, on difficult samples, it achieves the highest R-L (45.83) and BERTScore (89.09). On the easy subset, it maintains a clear lead with BLEU 57.01, R-L 63.08, and BERTScore 93.05. Similarly, Figure~\ref{fig:easy+diff} shows that the performance profile of Llama-3.1-8B-FT (EPD) surpasses other fine-tuned models across both subsets. These results confirm the robustness of Llama-3.1-8B-FT (EPD), which excels not only on easier cases but also in structurally challenging scenarios. This trend also holds for few-shot settings, where the model with one-shot example from EPO outperforms its counterparts on both difficult and easy samples.

\noindent \textbf{All models degrade significantly on difficult samples.}
As expected, all models exhibit a marked performance drop on the difficult subset of EPD. For example, as shown in Table~\ref{table:rq2_main2}, Llama-3.1-8B's BLEU score drops from 37.14 on easy samples to 24.41 on difficult ones, with similar trends observed across other metrics and models.

\begin{figure}[!t]
    \centering   
    \includegraphics[width=.49\textwidth]{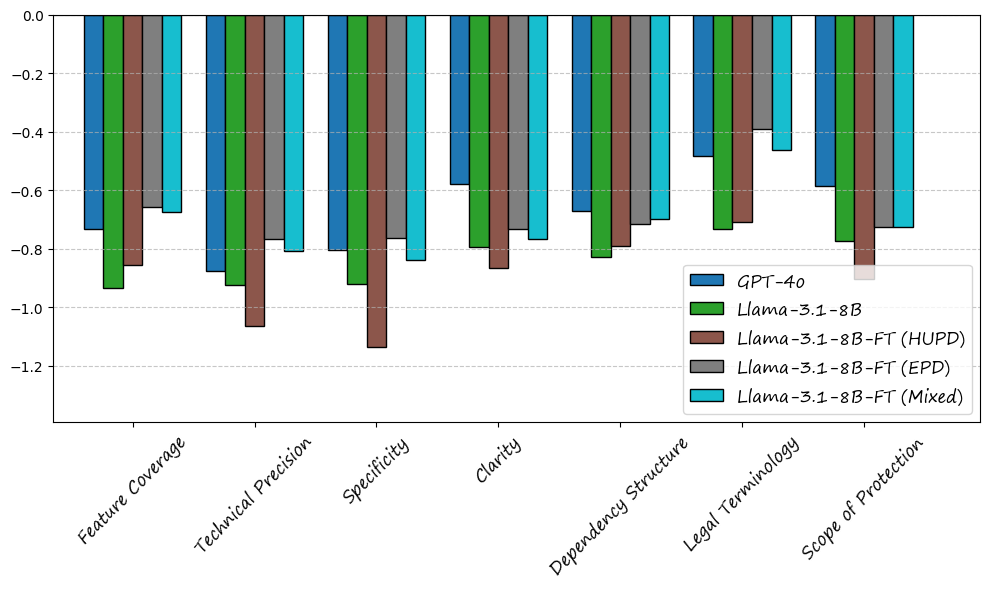}   
    \caption{LLM-as-a-judge performance difference between difficult and easy subsets (difficult minus easy).}
    \label{fig:diff-easy}
\end{figure}

Figure~\ref{fig:diff-easy} further illustrates the degradation across various LLM-as-a-judge dimensions. Llama-3.1-8B-FT (HUPD-DCG) is most affected, with declines exceeding 1.0 in specificity and precision. Even strong models like GPT-4o and Llama-3.1-8B-FT (EPD) show consistent drops of 0.4–0.8 across most dimensions.

These results highlight the intrinsic difficulty of these samples and the limitations of current models in handling them. While easy samples often allow for near-verbatim extraction from the description, difficult cases require the model to extract key details, rephrase, and reconstruct them into coherent and legally sound claims. The construction of the difficult subset within EPD thus provides a valuable benchmark for future research on model robustness, handling of complex dependencies, and generalization to low-frequency structures in legal text.

\subsection{Human Evaluation and Qualitative Analysis}
We provide example claims generated by different models on EPD in Appendix Table \ref{table:example_ouput}. We compare model outputs with gold claims and conduct a qualitative assessment to ensure the result analysis is self-contained.

Llama-3.1-8B omits key technical details present in the gold claim, such as \textit{the cartridge can be recharged} and \textit{manually activate}. Additionally, some claims exhibit redundancy, for example, both claims 2 and 7 mention \textit{detecting the user's physical condition}. The model also introduces extra features not included in the gold claim, such as \textit{a depth sensor} and \textit{emergency services}. 

Fine-tuned models demonstrate improvements in multiple aspects, such as feature completeness, structural coherence, and language precision. They successfully retain key features like \textit{rechargeable} and \textit{manually activate}, while reducing redundancy. Compared to Llama-3.1-8B-FT (EPD), Llama-3.1-8B-FT (HUPD-DCG) introduces an additional claim (claim 7) that describes alternative gas generation methods (compressed gas), which is not in the gold claim. This example reflects drafting convention differences between USPTO and EPO, where USPTO favors broader claim coverage through added variations, while EPO emphasizes precision and clarity.

The mixed model further illustrates these contrasts across jurisdictions. It uses the more generic term \textit{cartridge of compressed gas} rather than the more specific \textit{cartridge filled with CO2} used in the gold and other fine-tuned models. It also simplifies some phrasing, such as replacing \textit{said inflatable collar} with \textit{said collar}. Moreover, claims 4, 5, and 6 generated by Llama-3.1-8B-FT (EPD) directly reference claim 1, whereas the mixed model uses \textit{according to any of the previous claims}. These patterns suggest the mixed model tends to prioritize more general combinations of claims, which, as multiple dependent claims, could incur additional fees at the EPO.

GPT-4o produces well-structured and fluent claims but omits key elements such as \textit{emission/reception means} and \textit{predetermined values} included in the gold claim and fine-tuned models. It also adds extra features that expand beyond the core inventive concept, such as alternative gas generation methods, suitability for various user groups, and external alert systems. This suggests GPT-4o is pre-trained toward the US drafting style, likely influenced by its extensive training on USPTO data.

Overall, the model fine-tuned on EPD aligns most closely with the gold claim in terms of technical features, structure, and scope. It demonstrates the effectiveness of the EPD dataset in guiding high-quality claim generation.

\section{Conclusion}

We present the EPD dataset, a collection of European patents to support various patent-related tasks.  Compared to existing datasets for patent claim generation, EPD offers three major advantages: higher quality, closer to real-world situations, and lower risk of data leakage. EPD fills a key gap for cross-jurisdiction evaluation of LLMs on patent tasks. 
We conduct extensive experiments on both the HUPD-DCG dataset (from USPTO) and our EPD. Results demonstrate that models fine-tuned on EPD consistently outperform those trained on HUPD-DCG and even GPT-4o across both test sets, which highlights EPD's effectiveness in enhancing claim generation quality and cross-domain generalization. Additionally, we introduce a difficult subset of EPD to simulate real-world claim drafting challenges. Our analysis shows that all tested models experience a substantial performance drop on these samples.
Overall, our findings provide valuable insights into the limitations and opportunities in automated patent claim generation. We hope EPD will serve as a foundation for future research toward stronger claim-generation systems.

\section*{Limitations}
The EPD dataset only includes patents published by the European Patent Office and documented in English. For claim generation, we use patent descriptions constrained to fewer than 8,000 tokens to ensure a fair comparison with previous research. Additionally, we do not perform hyperparameter tuning during fine-tuning or inference. 

\section*{Ethics Statement}
Llama-3 is under \textit{META LLAMA 3 COMMUNITY LICENSE AGREEMENT}. GPT-4o is under a commercial license provided by OpenAI, and we access it through its API. Our dataset is collected from EPO's Open Patent Services (OPS). According to the rule of \textit{Terms and Conditions for use of the EPO's OPS}, we will provide a list of publication numbers and code to enable users to easily obtain structured data directly from OPS. Our proposed dataset is organized for multiple patent-related tasks and is compatible with the original access and use conditions. We plan to use the \textit{CC-BY-NC-4.0} license. Although the dataset contains the inventors' information on each patent, these details are already in the public domain and available for retrieval on EPO systems. The use of existing artifacts is consistent with their intended use.

\bibliography{anthology,custom}

\begin{thebibliography}{38}
\providecommand{\natexlab}[1]{#1}

\bibitem[{Bai et~al.(2024)Bai, Zhang, Chen, Cai, Zhong, Wang, Fang, Fang, Sun, Wang et~al.}]{bai2024patentgpt}
Zilong Bai, Ruiji Zhang, Linqing Chen, Qijun Cai, Yuan Zhong, Cong Wang, Yan Fang, Jie Fang, Jing Sun, Weikuan Wang, et~al. 2024.
\newblock Patentgpt: A large language model for intellectual property.
\newblock \emph{arXiv preprint arXiv:2404.18255}.

\bibitem[{Colombo et~al.(2024)Colombo, Pires, Boudiaf, Culver, Melo, Corro, Martins, Esposito, Raposo, Morgado et~al.}]{colombo2024saullm}
Pierre Colombo, Telmo~Pessoa Pires, Malik Boudiaf, Dominic Culver, Rui Melo, Caio Corro, Andre~FT Martins, Fabrizio Esposito, Vera~L{\'u}cia Raposo, Sofia Morgado, et~al. 2024.
\newblock Saullm-7b: A pioneering large language model for law.
\newblock \emph{arXiv preprint arXiv:2403.03883}.

\bibitem[{DeepSeek-AI(2024)}]{deepseekai2024deepseekv3technicalreport}
DeepSeek-AI. 2024.
\newblock \href {https://arxiv.org/abs/2412.19437} {Deepseek-v3 technical report}.
\newblock \emph{Preprint}, arXiv:2412.19437.

\bibitem[{Dubey et~al.(2024)Dubey, Jauhri, Pandey, Kadian, Al-Dahle, Letman, Mathur, Schelten, Yang, Fan et~al.}]{dubey2024llama}
Abhimanyu Dubey, Abhinav Jauhri, Abhinav Pandey, Abhishek Kadian, Ahmad Al-Dahle, Aiesha Letman, Akhil Mathur, Alan Schelten, Amy Yang, Angela Fan, et~al. 2024.
\newblock The llama 3 herd of models.
\newblock \emph{arXiv preprint arXiv:2407.21783}.

\bibitem[{{European Patent Office}(2000)}]{epo2020}
{European Patent Office}. 2000.
\newblock {EPC -- The European Patent Convention}.
\newblock \url{https://www.epo.org/en/legal/epc/2020/regulations.html}.
\newblock Accessed: 2023-06-12.

\bibitem[{Fall et~al.(2003)Fall, T{\"o}rcsv{\'a}ri, Benzineb, and Karetka}]{fall2003automated}
Caspar~J Fall, Atilla T{\"o}rcsv{\'a}ri, Karim Benzineb, and Gabor Karetka. 2003.
\newblock Automated categorization in the international patent classification.
\newblock In \emph{Acm Sigir Forum}, volume~37, pages 10--25. ACM New York, NY, USA.

\bibitem[{Frumkin(1947)}]{Frumkin1947}
M.~Frumkin. 1947.
\newblock Early history of patents for innovation.
\newblock \emph{Transactions of the Newcomen Society}, 26(1):47--56.

\bibitem[{Hu et~al.(2021)Hu, Wallis, Allen-Zhu, Li, Wang, Wang, Chen et~al.}]{hu2021lora}
Edward~J Hu, Phillip Wallis, Zeyuan Allen-Zhu, Yuanzhi Li, Shean Wang, Lu~Wang, Weizhu Chen, et~al. 2021.
\newblock Lora: Low-rank adaptation of large language models.
\newblock In \emph{International Conference on Learning Representations}.

\bibitem[{Jiang and Goetz(2025)}]{jiang2025natural}
Lekang Jiang and Stephan~M Goetz. 2025.
\newblock Natural language processing in the patent domain: a survey.
\newblock \emph{Artificial Intelligence Review}, 58(7):214.

\bibitem[{Jiang et~al.(2025{\natexlab{a}})Jiang, Scherz, and Goetz}]{jiang-etal-2025-patent}
Lekang Jiang, Pascal~A. Scherz, and Stefan Goetz. 2025{\natexlab{a}}.
\newblock \href {https://doi.org/10.18653/v1/2025.naacl-long.116} {Patent-{CR}: A dataset for patent claim revision}.
\newblock In \emph{Proceedings of the 2025 Conference of the Nations of the Americas Chapter of the Association for Computational Linguistics: Human Language Technologies (Volume 1: Long Papers)}, pages 2300--2314, Albuquerque, New Mexico. Association for Computational Linguistics.

\bibitem[{Jiang et~al.(2025{\natexlab{b}})Jiang, Scherz, and Goetz}]{jiang-etal-2025-towards-better}
Lekang Jiang, Pascal~A. Scherz, and Stefan Goetz. 2025{\natexlab{b}}.
\newblock \href {https://doi.org/10.18653/v1/2025.acl-long.190} {Towards better evaluation for generated patent claims}.
\newblock In \emph{Proceedings of the 63rd Annual Meeting of the Association for Computational Linguistics (Volume 1: Long Papers)}, pages 3775--3788, Vienna, Austria. Association for Computational Linguistics.

\bibitem[{Jiang et~al.(2026{\natexlab{a}})Jiang, Sun, and Goetz}]{jiang-etal-2026-reasoning}
Lekang Jiang, Wenjun Sun, and Stefan Goetz. 2026{\natexlab{a}}.
\newblock \href {https://doi.org/10.18653/v1/2026.findings-acl.541} {Reasoning for hierarchical text classification: The case of patents}.
\newblock In \emph{Findings of the {A}ssociation for {C}omputational {L}inguistics: {ACL} 2026}, pages 11127--11142, San Diego, California, United States. Association for Computational Linguistics.

\bibitem[{Jiang et~al.(2026{\natexlab{b}})Jiang, Sun, and Goetz}]{jiang2026benchmarking}
Lekang Jiang, Wenjun Sun, and Stephan Goetz. 2026{\natexlab{b}}.
\newblock Benchmarking patent drafting from inventor-style disclosures.
\newblock \emph{arXiv preprint arXiv:2608.21249}.

\bibitem[{Jiang et~al.(2026{\natexlab{c}})Jiang, Sun, and Goetz}]{jiang2026reasoning}
Lekang Jiang, Wenjun Sun, and Stephan Goetz. 2026{\natexlab{c}}.
\newblock \href {https://arxiv.org/abs/2607.10480} {When reasoning hurts legal drafting: The verbalization bottleneck in patent claim generation}.
\newblock \emph{arXiv preprint arXiv:2607.10480}.

\bibitem[{Jiang et~al.(2025{\natexlab{c}})Jiang, Zhang, Scherz, and Goetz}]{jiang-etal-2025-large}
Lekang Jiang, Caiqi Zhang, Pascal~A. Scherz, and Stefan Goetz. 2025{\natexlab{c}}.
\newblock \href {https://doi.org/10.18653/v1/2025.findings-naacl.70} {Can large language models generate high-quality patent claims?}
\newblock In \emph{Findings of the Association for Computational Linguistics: NAACL 2025}, pages 1272--1287, Albuquerque, New Mexico. Association for Computational Linguistics.

\bibitem[{Kincaid et~al.(1975)Kincaid, Fishburne~Jr, Rogers, and Chissom}]{kincaid1975derivation}
J~Peter Kincaid, Robert~P Fishburne~Jr, Richard~L Rogers, and Brad~S Chissom. 1975.
\newblock Derivation of new readability formulas (automated readability index, fog count and flesch reading ease formula) for navy enlisted personnel.
\newblock \emph{Technical report, Naval Technical Training Command Millington TN Research Branch}.

\bibitem[{Lee(2020)}]{lee2020controlling}
Jieh-Sheng Lee. 2020.
\newblock Controlling patent text generation by structural metadata.
\newblock In \emph{Proceedings of the 29th ACM International Conference on Information \& Knowledge Management}, pages 3241--3244.

\bibitem[{Lee(2023)}]{lee2023evaluating}
Jieh-Sheng Lee. 2023.
\newblock Evaluating generative patent language models.
\newblock \emph{World Patent Information}, 72:102173.

\bibitem[{Lee and Hsiang(2020)}]{lee2020patentgenerate}
Jieh-Sheng Lee and Jieh Hsiang. 2020.
\newblock Patent claim generation by fine-tuning openai gpt-2.
\newblock \emph{World Patent Information}, 62:101983.

\bibitem[{Li et~al.(2025)Li, Zhou, Glava{\v{s}}, Korhonen, and Vuli{\'c}}]{li-etal-2025-large-language-models}
Chengzu Li, Han Zhou, Goran Glava{\v{s}}, Anna Korhonen, and Ivan Vuli{\'c}. 2025.
\newblock \href {https://doi.org/10.18653/v1/2025.findings-acl.603} {Large language models are miscalibrated in-context learners}.
\newblock In \emph{Findings of the Association for Computational Linguistics: ACL 2025}, pages 11575--11596, Vienna, Austria. Association for Computational Linguistics.

\bibitem[{Li et~al.(2018)Li, Hu, Cui, and Hu}]{li2018deeppatent}
Shaobo Li, Jie Hu, Yuxin Cui, and Jianjun Hu. 2018.
\newblock Deeppatent: patent classification with convolutional neural networks and word embedding.
\newblock \emph{Scientometrics}, 117:721--744.

\bibitem[{Lin(2004)}]{lin2004rouge}
Chin-Yew Lin. 2004.
\newblock Rouge: A package for automatic evaluation of summaries.
\newblock In \emph{Text summarization branches out}, pages 74--81.

\bibitem[{Liu et~al.(2024)Liu, Feng, Xue, Wang, Wu, Lu, Zhao, Deng, Zhang, Ruan et~al.}]{liu2024deepseek}
Aixin Liu, Bei Feng, Bing Xue, Bingxuan Wang, Bochao Wu, Chengda Lu, Chenggang Zhao, Chengqi Deng, Chenyu Zhang, Chong Ruan, et~al. 2024.
\newblock Deepseek-v3 technical report.
\newblock \emph{arXiv preprint arXiv:2412.19437}.

\bibitem[{Liu et~al.(2023)Liu, Iter, Xu, Wang, Xu, and Zhu}]{liu-etal-2023-g}
Yang Liu, Dan Iter, Yichong Xu, Shuohang Wang, Ruochen Xu, and Chenguang Zhu. 2023.
\newblock \href {https://doi.org/10.18653/v1/2023.emnlp-main.153} {{G}-eval: {NLG} evaluation using gpt-4 with better human alignment}.
\newblock In \emph{Proceedings of the 2023 Conference on Empirical Methods in Natural Language Processing}, pages 2511--2522, Singapore. Association for Computational Linguistics.

\bibitem[{Ouyang et~al.(2022)Ouyang, Wu, Jiang, Almeida, Wainwright, Mishkin, Zhang, Agarwal, Slama, Ray et~al.}]{ouyang2022training}
Long Ouyang, Jeffrey Wu, Xu~Jiang, Diogo Almeida, Carroll Wainwright, Pamela Mishkin, Chong Zhang, Sandhini Agarwal, Katarina Slama, Alex Ray, et~al. 2022.
\newblock Training language models to follow instructions with human feedback.
\newblock \emph{Advances in neural information processing systems}, 35:27730--27744.

\bibitem[{Papineni et~al.(2002)Papineni, Roukos, Ward, and Zhu}]{papineni2002bleu}
Kishore Papineni, Salim Roukos, Todd Ward, and Wei-Jing Zhu. 2002.
\newblock Bleu: a method for automatic evaluation of machine translation.
\newblock In \emph{Proceedings of the 40th annual meeting of the Association for Computational Linguistics}, pages 311--318.

\bibitem[{Rafailov et~al.(2024)Rafailov, Sharma, Mitchell, Manning, Ermon, and Finn}]{rafailov2024direct}
Rafael Rafailov, Archit Sharma, Eric Mitchell, Christopher~D Manning, Stefano Ermon, and Chelsea Finn. 2024.
\newblock Direct preference optimization: Your language model is secretly a reward model.
\newblock \emph{Advances in Neural Information Processing Systems}, 36.

\bibitem[{Risch et~al.(2020)Risch, Alder, Hewel, and Krestel}]{risch2020patentmatch}
Julian Risch, Nicolas Alder, Christoph Hewel, and Ralf Krestel. 2020.
\newblock Patentmatch: a dataset for matching patent claims \& prior art.
\newblock \emph{arXiv preprint arXiv:2012.13919}.

\bibitem[{Schulman et~al.(2017)Schulman, Wolski, Dhariwal, Radford, and Klimov}]{schulman2017proximal}
John Schulman, Filip Wolski, Prafulla Dhariwal, Alec Radford, and Oleg Klimov. 2017.
\newblock Proximal policy optimization algorithms.
\newblock \emph{arXiv preprint arXiv:1707.06347}.

\bibitem[{Shalaby and Zadrozny(2019)}]{shalaby2019patent}
Walid Shalaby and Wlodek Zadrozny. 2019.
\newblock Patent retrieval: a literature review.
\newblock \emph{Knowledge and Information Systems}, 61:631--660.

\bibitem[{Sharma et~al.(2019)Sharma, Li, and Wang}]{sharma-etal-2019-bigpatent}
Eva Sharma, Chen Li, and Lu~Wang. 2019.
\newblock \href {https://doi.org/10.18653/v1/P19-1212} {{BIGPATENT}: A large-scale dataset for abstractive and coherent summarization}.
\newblock In \emph{Proceedings of the 57th Annual Meeting of the Association for Computational Linguistics}, pages 2204--2213, Florence, Italy. Association for Computational Linguistics.

\bibitem[{Sun et~al.(2020)Sun, Wang, Li, Feng, Tian, Wu, and Wang}]{sun2020ernie}
Yu~Sun, Shuohuan Wang, Yukun Li, Shikun Feng, Hao Tian, Hua Wu, and Haifeng Wang. 2020.
\newblock Ernie 2.0: A continual pre-training framework for language understanding.
\newblock In \emph{Proceedings of the AAAI conference on artificial intelligence}, volume~34, pages 8968--8975.

\bibitem[{Suzgun et~al.(2024)Suzgun, Melas-Kyriazi, Sarkar, Kominers, and Shieber}]{suzgun2024harvard}
Mirac Suzgun, Luke Melas-Kyriazi, Suproteem Sarkar, Scott~D Kominers, and Stuart Shieber. 2024.
\newblock The harvard uspto patent dataset: A large-scale, well-structured, and multi-purpose corpus of patent applications.
\newblock \emph{Advances in Neural Information Processing Systems}, 36.

\bibitem[{Wang et~al.(2024)Wang, Ni, Liu, Lu, Chen, Feng, Wei, Qu, Alinejad-Rokny, Lin et~al.}]{wang2024autopatent}
Qiyao Wang, Shiwen Ni, Huaren Liu, Shule Lu, Guhong Chen, Xi~Feng, Chi Wei, Qiang Qu, Hamid Alinejad-Rokny, Yuan Lin, et~al. 2024.
\newblock Autopatent: A multi-agent framework for automatic patent generation.
\newblock \emph{arXiv preprint arXiv:2412.09796}.

\bibitem[{Wei et~al.(2022)Wei, Wang, Schuurmans, Bosma, Xia, Chi, Le, Zhou et~al.}]{wei2022chain}
Jason Wei, Xuezhi Wang, Dale Schuurmans, Maarten Bosma, Fei Xia, Ed~Chi, Quoc~V Le, Denny Zhou, et~al. 2022.
\newblock Chain-of-thought prompting elicits reasoning in large language models.
\newblock \emph{Advances in neural information processing systems}, 35:24824--24837.

\bibitem[{Zhang et~al.(2019)Zhang, Kishore, Wu, Weinberger, and Artzi}]{zhang2019bertscore}
Tianyi Zhang, Varsha Kishore, Felix Wu, Kilian~Q Weinberger, and Yoav Artzi. 2019.
\newblock Bertscore: Evaluating text generation with bert.
\newblock In \emph{International Conference on Learning Representations}.

\bibitem[{Zhao et~al.(2023)Zhao, Zhou, Li, Tang, Wang, Hou, Min, Zhang, Zhang, Dong et~al.}]{zhao2023survey}
Wayne~Xin Zhao, Kun Zhou, Junyi Li, Tianyi Tang, Xiaolei Wang, Yupeng Hou, Yingqian Min, Beichen Zhang, Junjie Zhang, Zican Dong, et~al. 2023.
\newblock A survey of large language models.
\newblock \emph{arXiv preprint arXiv:2303.18223}.

\bibitem[{Zheng et~al.(2024)Zheng, Zhang, Zhang, Ye, and Luo}]{zheng-etal-2024-llamafactory}
Yaowei Zheng, Richong Zhang, Junhao Zhang, Yanhan Ye, and Zheyan Luo. 2024.
\newblock \href {https://doi.org/10.18653/v1/2024.acl-demos.38} {{L}lama{F}actory: Unified efficient fine-tuning of 100+ language models}.
\newblock In \emph{Proceedings of the 62nd Annual Meeting of the Association for Computational Linguistics (Volume 3: System Demonstrations)}, pages 400--410, Bangkok, Thailand. Association for Computational Linguistics.

\end{thebibliography}

\appendix

\begin{table*}[!ht]
\footnotesize
\centering

\begin{tabular}{l|l|l}
\toprule
\textbf{Structure} & \textbf{Label}      & \multicolumn{1}{c}{\textbf{Description}}   \\ 
\midrule
Section   & F          & Mechanical engineering ; Lighting; Heating; Weapons; Blasting  \\
Class     & F02        & Combustion engines; hot-gas or combustion-product engine plants \\
Sub-class & F02D       & Controlling combustion engines         \\
Group     & F02D 41    & Electrical control of supply of combustible mixture of its constituents \\
Sub-group & F02D 41/02 & Circuit arrangements for generating control signals \\ 
\bottomrule
\end{tabular}
\caption{Example of International Patent Classification (IPC) scheme from \citet{jiang2025natural}}
\label{ipc_scheme}
\end{table*}
\begin{table*}[ht!]
\centering
\footnotesize
\begin{tabular}{l|l|c|c}
\toprule
\textbf{Section} & \multicolumn{1}{c|}{\textbf{Description}}         & \textbf{Single} & \textbf{Both} \\ \midrule                                             
A &  Human necessities & 11,152 & 6,223 \\
B &  Performing operations; transporting & 9,754 & 7,003 \\
C &  Chemistry; metallurgy & 3,308 & 2,249 \\
D &  Textiles; paper & 393 & 326 \\
E &  Fixed constructions & 1,683 & 753 \\
F &  Mechanical engineering; lighting; heating; weapons; blasting & 3,260 & 2,973 \\
G &  Physics & 6,897 & 6,963 \\
H &  Electricity & 5,414 & 6,713 \\\bottomrule
\end{tabular}
\caption{Number of documents in different IPC sections }
\label{table:dataset_sections}
\end{table*}

\section{Patent Background}
\label{background}

Patent documents are essential for protecting intellectual property (IP) and documenting inventions. They provide a detailed description of new inventions and define the scope of patent rights granted to the holder. As integral components of the patenting process, these documents become publicly accessible once a patent is issued. Although the format and content may vary by jurisdiction, they typically include the following elements: title, bibliometric information, patent classification code, citations, abstract, drawings, detailed description, and claims. \citet{jiang2025natural} identified three key challenges for LLMs in patent-related tasks: handling the long context length, processing technical language, and satisfying precision requirements. Researchers showed the difficulties of applying LLMs to patent-related tasks \citep{jiang-etal-2025-patent,jiang-etal-2025-towards-better,jiang-etal-2025-large,jiang-etal-2026-reasoning,jiang2026benchmarking,jiang2026reasoning}.

Patents are organized into hierarchical categories to facilitate efficient searching and management. The International Patent Classification  (IPC) and Cooperative Patent Classification (CPC) systems are two of the most popular classification schemes. For example, the IPC system classifies patents into six levels: section, class, subclass, group, and subgroup. Table \ref{ipc_scheme} lists the breakdown of the F02D 41/02 label using the IPC scheme.

\section{Dataset Details}
\label{datadetails}

\textbf{Data Structure and Format}
Each patent document is saved to a structured JSON file, named by its publication number and version, such as EP1234567B1.
The data fields include publication numbers, title, kind code, publication dates, primary and secondary classification codes, patent family, application reference, applicants, inventors, citations, abstract, claims, and descriptions. 
 
\noindent \textbf{More Statistics} Table \ref{table:dataset_sections} illustrates the number of documents across different IPC sections in this dataset. 

\begin{figure*}[t!]
\begin{lstlisting}[language=json, xleftmargin=2em, xrightmargin=2em]
{
  "publication_number": "...",
  "title": "...",
  "kind_code": "...",
  "date_published": "...",
  "main_cpci_label": "...",
  "cpci_labels": ["...", "...", "..."],
  "main_ipcr_label": "...",
  "ipcr_labels": ["...", "...", "..."],
  "family_id": "...",
  "application_reference": "...",
  "applicants": ["...", "...", "..."],
  "inventors": ["...", "...", "..."],
  "citations": [
        {
            "cited_phase": "...",
            "cited_by": "...",
            "source": "...",
            "text": "...",
            "document_number": "...",
            "category": "...",
            "related_claims": "...",
            "related_passages": ["...", "...", "..."]
        },
        {...},
        ...
   ]
  "abstract": "...",
  "claims": "...",
  "description": "..."
}
\end{lstlisting}

\caption{An example of an organized data file in EPD}
\label{fig:exampledata}
\end{figure*}

\section{More Tasks}
\label{moretask}

Our EPD dataset supports multiple patent-related tasks. We highlight some other possible uses in this section. All tasks use patents from August 2024 as the test set, while the remaining patents serve as the training set. We select the Llama-3.1-8B-Instruct\footnote{\url{https://huggingface.co/meta-llama/Llama-3.1-8B-Instruct}} as the base model for all tasks because of its strong performance and a suitable size for fine-tuning. Details of experimental settings are provided in Appendix~\ref{expdetails}.

\subsection{Patent Subject Classification}

\textbf{Task definition }
This task involves classifying patents into predefined categories based on their subjects. Automating this classification process is crucial given the large number of patents being filed and the complexity of manual classification. 

\begin{table*}[!ht]
\centering
\footnotesize
\resizebox{\textwidth}{!}{
\begin{tabular}{l|cc|cc|cc|cc|cc|cc}
\toprule
\multirow{4}{*}{\textbf{Model}} 
& \multicolumn{6}{c|}{\textbf{HUPD}} 
& \multicolumn{6}{c}{\textbf{EPD}} \\ \cmidrule{2-13}
& \multicolumn{2}{c|}{\textbf{Section}} 
& \multicolumn{2}{c|}{\textbf{Class}} 
& \multicolumn{2}{c|}{\textbf{Subclass}} 
& \multicolumn{2}{c|}{\textbf{Section}} 
& \multicolumn{2}{c|}{\textbf{Class}} 
& \multicolumn{2}{c}{\textbf{Subclass}} \\ \cmidrule{2-13}
& \textbf{Acc.} & \textbf{F1} 
& \textbf{Acc.} & \textbf{F1} 
& \textbf{Acc.} & \textbf{F1} 
& \textbf{Acc.} & \textbf{F1} 
& \textbf{Acc.} & \textbf{F1} 
& \textbf{Acc.} & \textbf{F1} \\ \midrule
Llama-3.1-8B 
& 56.37 & 36.03 & 32.64 & 17.11 & 25.48 & 12.11 
& 51.59 & 41.02 & 31.21 & 17.59 & 21.64 & 10.37 \\
Llama-3.1-8B-FT (HUPD) 
& 81.69 & 74.23 & 72.84 & 50.98 & 63.54 & 36.27 
& 76.13 & 71.42 & 61.74 & 42.96 & 47.34 & 27.36 \\
Llama-3.1-8B-FT (EPD) 
& 78.41 & 71.73 & 67.12 & 45.94 & 55.22 & 30.51 
& 80.77 & 74.77 & 67.63 & 48.44 & 54.78 & 33.41 \\
\bottomrule
\end{tabular}
}
\caption{Patent classification results on HUPD and EPD datasets (Accuracy and F1).}
\label{table:patent_class_results}
\end{table*}

Since there are significant overlaps between IPC and CPC, we use IPC codes as labels for consistency with prior work \citep{suzgun2024harvard}. We focus on three levels of prediction: section, class, and subclass. We explore the model performance under two different settings similar to previous work \citep{suzgun2024harvard}: fine-tuning based on abstracts.

\noindent \textbf{Results }
Table \ref{table:patent_class_results} shows the results of this multi-class subject classification task based on the IPC labels at section, class, and subclass levels. We observe that models fine-tuned on one jurisdiction (e.g., USPTO vs. EPO) perform slightly worse on the other. This drop is due to differences in examiner practices, technical focus, and the interpretation of “technical contribution”, which leads to divergent labeling standards and thus different learning objectives during fine-tuning.

\subsection{Summarization}

\begin{table*}[!ht]
\centering
\footnotesize
\begin{tabular}{l|ccc|ccc}
\toprule
\multirow{2.7}{*}{\textbf{Model}} & \multicolumn{3}{c|}{\textbf{HUPD}} & \multicolumn{3}{c}{\textbf{EPD}} \\ \cmidrule{2-7}
 & \textbf{R-1} & \textbf{R-2} & \textbf{R-L}  & \textbf{R-1} & \textbf{R-2} & \textbf{R-L}\\ \midrule
Llama-3.1-8B & 53.49 & 31.65 & 39.06 & 56.14 & 36.08 & 42.96 \\
Llama-3.1-8B-FT (HUPD) & 60.88 & 44.43 & 50.56 & 69.66 & 56.28 & 62.08 \\
Llama-3.1-8B-FT (EPD) & 59.71 & 43.70 & 50.10 & 69.48 & 56.77 & 62.60 \\

\bottomrule
\end{tabular}
\caption{Results of patent summarization as measured by ROUGE score.}
\label{table:patent_sum_results}
\end{table*}

\textbf{Task definition }
Patent summarization aims to generate concise yet informative abstracts from patent claims or descriptions. It uses more accessible language to briefly introduce the complex and technical patent to wider audiences. This process involves identifying and condensing the key aspects of a patent, such as claims and detailed descriptions of the invention.

Previous research showed that claim-based abstract generation outperformed description-based summarization \citep{suzgun2024harvard}. Hence, we focus on generating patent abstracts based on the claims. 

\noindent \textbf{Results }
Table \ref{table:patent_sum_results} provides results of patent summarization as measured by ROUGE score \citep{lin2004rouge}. We find that the performance of fine-tuned models remains relatively consistent across jurisdictions. This is likely because both USPTO and EPO adhere to similar standards for abstract writing, which makes the task more domain-invariant.

\subsection{Language Modeling }
This dataset provides approximately 0.12 billion tokens of patent texts in 2024, which is not likely to appear in the pre-training dataset of some current LLMs. Therefore, this dataset supports two language modeling tasks. First, researchers can do domain-specific training to adapt general LLMs to the patent or legal domain \citep{colombo2024saullm}. Since patent tasks are particularly challenging due to the complex terminology, extensive length, and technical contents, patent-specific LLMs are very promising to improve performance. Second, researchers can do continual pre-training \citep{sun2020ernie} to keep current patent-specific models up-to-date. Such models will be more adaptive for analyzing or managing recently published patents. Extensive high-quality and recent data is helpful to both tasks. 

\subsection{Patent Retrieval }
Patent retrieval aims to retrieve patent documents related to a given patent \citep{shalaby2019patent}. This process is crucial for patent examiners to assess the patentability of a new patent application. Since the dataset includes patent citations, researchers could easily formulate pairs of patents relevant to each other. 

\section{More Results}

\begin{table*}[ht!]
\footnotesize
\centering
\resizebox{\linewidth}{!}{
\begin{tabular}{lcccccccccc}
\toprule
& \multicolumn{4}{c}{\textbf{HUPD-DCG}} & \multicolumn{4}{c}{\textbf{EPD}}  \\
\cmidrule(lr){2-5} \cmidrule(lr){6-9}
\textbf{Model} & \textbf{BLEU} & \textbf{R-1} & \textbf{R-L} & \textbf{BERTScore} & \textbf{BLEU} & \textbf{R-1} & \textbf{R-L} & \textbf{BERTScore} \\
\midrule

\multicolumn{3}{l}{\textbf{Fine-tuned LLMs with RLHF}} \\
Llama-3.1-8B-FT (DPO) & 20.85 & 51.88 & 31.45 & 84.10 & 33.05 & 57.69 & 34.14 & 86.13 \\
Llama-3.1-8B-FT (Mixed-DPO) & 25.96 & 53.69 & 36.14 & 85.83 & 38.41  & 59.89 & 38.44 & 87.45 \\

\bottomrule
\end{tabular}
}
\caption{Claim generation results on HUPD-DCG (from USPTO) and EPD (from EPO) datasets. }
\label{table:more1}
\end{table*}

\begin{table*}[ht!]
\footnotesize
\centering
\resizebox{\linewidth}{!}{
\begin{tabular}{lcccccccccc}
\toprule
& \multicolumn{4}{c}{\textbf{Difficult Samples}} & \multicolumn{4}{c}{\textbf{Easy Samples}}  \\
\cmidrule(lr){2-5} \cmidrule(lr){6-9}
\textbf{Model} & \textbf{BLEU} & \textbf{R-1} & \textbf{R-L} & \textbf{BERTScore} & \textbf{BLEU} & \textbf{R-1} & \textbf{R-L} & \textbf{BERTScore} \\
\midrule

\multicolumn{3}{l}{\textbf{Fine-tuned LLMs with RLHF}} \\
Llama-3.1-8B-FT (DPO) & 31.51 & 56.99 & 32.43 & 85.83 & 36.10 & 59.03 & 37.61 & 86.73 \\
Llama-3.1-8B-FT (Mixed-DPO) & 35.74 & 58.39 & 35.72 & 86.81 & 43.71 & 62.64 & 43.77 & 88.74 \\

\bottomrule
\end{tabular}
}
\caption{Claim generation results on difficult and easy examples of the EPD dataset.  }
\label{table:more2}
\end{table*}

\subsection{Dataset}
The EPD dataset contains two versions of each patent: the application version, initially rejected by the examiner, and the granted version, which is the final version that passes the examination. These two versions naturally provide human-annotated quality labels, which can serve to enhance LLM alignment in patent claim generation, particularly through reinforcement learning with human feedback (RLHF) \citep{ouyang2022training}. We obtain 10,087 pairs of claims by filtering 32,988 patents from EPD that contain both versions and have description lengths under 8,000 tokens. Notably, there is no overlap between the preference dataset and the fine-tuning dataset.

\subsection{Models}
For the RLHF process, we use Direct Preference Optimization (DPO) \citep{rafailov2024direct} based on the preference dataset. Traditional RLHF methods, such as Proximal Policy Optimization (PPO) \citep{schulman2017proximal}, have been widely used in instruction tuning \citep{ouyang2022training}. However, PPO-based approaches require explicit reward modeling, are computationally expensive, and often lead to instability. DPO provides a more efficient alternative by directly optimizing model preferences without explicit reward modeling. Given these advantages, we apply DPO for the RLHF process in patent claim generation.
We train \textbf{Llama-3.1-8B-FT (DPO)} starting from the base Llama-3.1-8B model, while the \textbf{Llama-3.1-8B-FT (Mixed-DPO)} is trained from Llama-3.1-8B-FT (Mixed). We introduce experimental details in Appendix~\ref{expdetails}.

\subsection{Results}
Table \ref{table:more1} shows the performance of fine-tuned models with RLHF using DPO on both HUPD-DCG and EPD datasets. The results indicate that while fine-tuning significantly enhances claim generation performance over base models, DPO-based RLHF does not lead to improvements across all evaluation metrics. For example, on the HUPD-DCG dataset, the Llama-3.1-8B-FT (DPO) model achieves lower scores (BLEU: 20.85, R-L: 31.45, BERTScore: 84.10) compared to its purely fine-tuned counterpart (BLEU: 40.54, R-L: 48.70, BERTScore: 89.89). Interestingly, the DPO model even underperforms the original base model without fine-tuning (BLEU: 34.30, R-L: 40.91, BERTScore: 87.64). A similar trend is observed on the EPD dataset, which further confirms that DPO-based RLHF does not enhance claim generation quality.

We also find that applying DPO on a fine-tuned model leads to better results than applying DPO directly on a base model. However, its performance remains significantly below that of the fine-tuned model. This suggests that fine-tuned models already establish a well-formed distribution, and DPO optimization is not well-aligned with the claim generation objective.

These findings suggest that while the application-to-granted claim transformations provide a natural source of preference data, their direct application in DPO-based RLHF does not necessarily contribute to meaningful improvements in claim generation. Additional refinements or processing steps are required to construct an effective preference dataset for RLHF.

A similar trend is observed when evaluating both difficult and easy samples, as shown in Table \ref{table:more2}. In both datasets, LLMs trained with DPO perform even worse than the base model without fine-tuning.

\subsection{Analysis and Future Work}
A key observation from our experiments is that the rewarding accuracy achieves exceptionally high scores of over 98\% during the DPO process. Given that assessing the quality of patent claims is highly complex, even for experienced patent professionals, this accuracy level seems implausible. The high score suggests that the model may be over-fitting to strongly correlated superficial patterns rather than learning meaningful quality distinctions.

One possible issue is that the model learns to differentiate claims based on simple heuristics, such as claim length or the presence of certain legal terms (e.g., ‘wherein’), rather than deeper technical and legal aspects. Additionally, it may prioritize format consistency over actual content quality, which is problematic for real-world claim generation, where the differences between high- and low-quality claims are often subtle. If the model over-fits these simplistic patterns, DPO optimization may not effectively guide claim generation toward higher-quality outputs, which ultimately leads to degraded performance.

Future research can explore alternative formulations of preference datasets. Potential directions include incorporating further human-annotated preferences or refining models to capture more meaningful quality differences. More complicated RLHF techniques may be investigated to better align the optimization process with the complexities of patent claim writing.

\subsection{Qualitative Analysis}
As shown in Table \ref{table:example_ouput}, for the DPO model, claim 1 is excessively long with overloaded details, and Claims 2--5 also exhibit significant redundancy. Additionally, the model frequently uses terms such as \textit{wherein} and \textit{configured to}. These tendencies align with previous analysis and suggest that the model may over-fit to simplistic patterns to assess claim quality. It may incorrectly associate longer first claims with higher quality and result in an overly lengthy claim 1. Moreover, DPO appears to prioritize certain terminological patterns, likely influenced by preferences in the training dataset.

\section{Experimental Details}
\label{expdetails}

All fine-tuning and inference processes are conducted on NVIDIA A100 GPUs. We use the LLaMA-Factory framework \citep{zheng-etal-2024-llamafactory} for model fine-tuning and inference. We opt LoRA \citep{hu2021lora} for fine-tuning to reduce computational costs while maintaining comparable performance. The total running time is about 630 hours. We do not explore the effects of few-shot prompting because we have fine-tuned models for all tasks, which perform stronger and more robustly. For classification evaluation, we use the \textit{sklearn} Python library to calculate accuracy, precision, recall, and F1 score. We use the Huggingface \textit{evaluate} library to measure the BLEU, ROUGE, and BERTScore for text generation tasks. 

\noindent \textbf{Claim Generation}
The following hyper-parameters are used during fine-tuning: LoRA rank: 8, LoRA alpha: 16, learning rate: 5e-5, batch size: 2, number of epochs: 3, validation ratio: 10\%, cut-off length: 10,240.
The following hyper-parameters are used during DPO: beta: 0.1, learning rate: 5e-5, batch size: 1, number of epochs: 2, validation ratio: 10\%. 
For inference, we use the following prompt: \textit{You are a patent expert. Given the following patent description, generate patent claims.} We set the maximum new tokens to 1024, temperature to 0.1, and top\_p to 0.95.

\noindent \textbf{Patent Subject Classification}
The following hyper-parameters are used during training: LoRA rank: 8, LoRA alpha: 16, learning rate: 5e-5, batch size: 8, number of epochs: 3, validation ratio: 10\%. The max input length is set to 256 for abstracts and 2,048 for claims. 
We use the following prompt instruction: \textit{You are a patent expert. You will be given abstracts/claims of a patent. Your task is to determine the patent's International Patent Classification code at the subclass level.}

\noindent \textbf{Patent Summarization }
The following hyper-parameters are used during training: LoRA rank: 8, LoRA alpha: 16, learning rate: 5e-5, batch size: 4, number of epochs: 3, validation ratio: 10\%, max input length: 2,048, max output length: 256.
We use the following prompt instruction: \textit{You are a patent expert. You will be given claims of a patent. Your task is to generate the patent abstract.} 

\section{Statistics Calculation Method}
\label{statscalculation}

We use the \textit{tiktoken} Python library with the GPT-4o tokenizer to count the number of tokens. The claim length is computed as the total token count divided by the number of claims. We use the \textit{spaCy} Python library to obtain structural complexity, determined by the ratio of subordinate clauses to the total number of sentences. Syntactic tree depth is computed using a constituency parser and measures the depth of the syntactic parse tree of the main independent claim. Higher values indicate more deeply nested grammatical structures. Readability is evaluated using the Flesch-Kincaid Grade Level formula \citep{kincaid1975derivation}, where a lower score indicates greater readability. The calculation is performed using the \textit{textstat} Python library. Lexical diversity, is calculated by MTLD (Measure of Textual Lexical Diversity), where higher values indicate that the vocabulary is more varied and less repetitive. Term density, calculated as the proportion of proper nouns and named entities, reflects the concentration of technical terminology.

\section{LLM-as-a-judge Settings}
\label{llmasjudge}
We use the prompt in Table \ref{tab:evalprompt} to obtain LLM-as-a-judge evaluation scores. During inference, we set the temperature to 0.1 and top\_p to 0.95.

\begin{table*}[ht]
\centering
\footnotesize
\begin{tabular}{p{15.5cm}}
\toprule
\textbf{Gold claims} 1. Collar life preserver for water sports that includes a compressed inflatable life preserver (1), a cartridge (2) that is filled with CO2, a solenoid valve (3) in connection with the cartridge (2), a discharge valve (4), a sensor (5) associated with a microprocessor (6) programmed with an algorithm and emission/reception means between the valve (3) and the microprocessor (6), where the cartridge (2) can be recharged after each use, the algorithm constantly receives information of the user's body functions and by means of predetermined values knows when the user is unconscious, the microprocessor (6) sends a signal to the solenoid valve (3) of the CO2 cartridge (2) for its automatic filling and there is a cord to manually activate the CO2 cartridge; wherein the sensor (5) includes a cardiac sensor that detects an altered heart rate, characterized in that the sensor (5) further includesa pulse oximeter that detects a low level of oxygen saturation in the blood,a movement sensor and a sensor that captures muscle contractions. \\
\midrule

\textbf{Llama-3.1-8B} Claim 1: A device for automatic inflation of a life preserver, comprising:a collar placed around the user's neck;an inflatable life preserver (1) internally compressed, said inflatable life preserver being connected to a CO2 cartridge (2);a solenoid valve (3) connected to the CO2 cartridge (2) for filling the inflatable life preserver (1);a discharge valve (4) for emptying the inflatable life preserver (1) after use;a sensor (5) associated with a microprocessor (6) programmed with an algorithm to detect the user's body functions and determine when the user is unconscious; emission/reception means between the valve (3) and the microprocessor (6) for sending a signal to the solenoid valve (3) for automatic filling of the inflatable life preserver (1) when the user is unconscious. Claim 2: The device of claim 1, wherein the sensor (5) includes a pulse oximeter, a cardiac sensor, a muscle contraction sensor, a movement sensor, and a depth sensor for detecting the user's physical condition. Claim 3: The device of claim 1, wherein the sensor (5) is capable of sending external radio signals to be captured by life guards or emergency services when the user is unconscious. Claim 4: The device of claim 1, wherein the CO2 cartridge (2) can be replaced by any other means of gas generation, such as decompression of a solid by heat. Claim 5: The device of claim 1, further comprising a rechargeable battery for powering the electronic elements. Claim 6: The device of claim 1, wherein the inflatable life preserver (1) is designed to keep the user's head out of the water at all times, ensuring breathing can be maintained if the user is unconscious. Claim 7: The device of claim 1, wherein the device is designed for use in aquatic sports, such as surfing, and is capable of detecting the user's physical condition, including oxygen saturation, heart rate, and muscle contractions. Claim 8: The device of claim 1, wherein the device can be used as an alternative for other sports where there is no specific protection, such as underwater fishing or for children or persons with little mobility. Claim 9: The device of claim 1, wherein the device has a manual activation cord for the CO2 cartridge (2) in case of need. Claim 10: The device of claim 1, wherein the inflatable life preserver (1) has a discharge valve (4) for emptying the inflatable life preserver (1) after use. \\
\midrule
\textbf{Llama-3.1-8B-FT (EPO)} 1. Device for the protection of a user in aquatic sports, comprising a collar (1) that is placed around the user's neck, said collar (1) being inflatable by means of a cartridge (2) that is filled with CO2, said inflatable collar (1) comprising a solenoid valve (3) in connection with said cartridge (2) that allows its filling, said inflatable collar (1) also comprising a discharge valve (4) that permits emptying the inflatable collar (1) after its use, said inflatable collar (1) presenting a sensor (5) associated with a microprocessor (6) programmed with an algorithm that constantly receives information on the user's body functions and is capable, through predetermined values, of knowing when the user is unconscious, said device having emission/reception means between said valve (3) and said microprocessor (6), said sensor (5) being able to send a signal that alters the threshold of the predetermined value and said microprocessor (6) to send a signal to the solenoid valve (3) of the cartridge (2) of CO2 for the automatic filling, in such a way that this device keeps the user's head out of the water at the time in which the life preserver is used and, therefore, breathing can be maintained if the user is unconscious. 2. Device according to claim 1, characterised in that it comprises a cord to activate the cartridge of CO2 manually in case of need. 3. Device according to claim 1, characterised in that it comprises a rechargeable battery that keeps the electronic elements in operation. 4. Device according to claim 1, characterised in that said microprocessor (6), programmed with the predetermined algorithm, detects by means of said sensor (5) an altered signal of the information obtain and sends a signal for the opening of the solenoid valve (3) for the activation of the cartridge (2). 5. Device according to claim 1, characterised in that said sensor (5) includes a pulse oximeter that detects a low level of oxygen saturation in the blood, a cardiac sensor that detects an altered heart rate, a sensor that captures muscle contractions, a movement sensor, in addition to a depth sensor for athletes that practice water sports, in particular, surfing. 6. Device according to claim 1, characterised in that it comprises a sensor capable of sending external radio signals capable of being captured by life guards or emergency services when the unconsciousness of the user is detected. \\   
\midrule
\textbf{Llama-3.1-8B-FT (USPTO)} 1. A device for a life preserver, comprising: a collar composed internally by a compressed inflatable life preserver, the collar being placed around a user's neck and being inflated by means of a cartridge filled with CO2; the inflatable life preserver having a solenoid valve in connection with the cartridge that allows its filling; the inflatable collar having a discharge valve that permits emptying the inflatable collar after its use; the inflatable collar having a sensor associated with a microprocessor programmed with an algorithm that constantly receives information on the user's body functions and is capable, through predetermined values, of knowing when the user is unconscious; the device having emission/reception means between the valve and the microprocessor, with this sensor being able to send a signal that alters the threshold of the predetermined value and the microprocessor to send a signal to the solenoid valve of the cartridge of CO2 for the automatic filling, in such a way that this device keeps the user's head out of the water at the time in which the life preserver is used and, therefore, breathing can be maintained if the user is unconscious. \\

\textit{Continue on next page} \\

\end{tabular}

\end{table*}

\begin{table*}[ht]
\centering
\footnotesize
\begin{tabular}{p{15.5cm}}

2. The device of claim 1, wherein the device has a cord to activate the cartridge of CO2 manually in case of need. 3. The device of claim 1, wherein the device has a rechargeable battery that keeps the electronic elements in operation. 4. The device of claim 1, wherein the microprocessor, programmed with the predetermined algorithm, detects by means of the sensor an altered signal of the information obtain and sends a signal for the opening of the solenoid valve for the activation of the cartridge. 5. The device of claim 1, wherein the sensor includes a pulse oximeter that detects a low level of oxygen saturation in the blood, a cardiac sensor that detects an altered heart rate, a sensor that captures muscle contractions, a movement sensor, in addition to a depth sensor for athletes that practice water sports, in particular, surfing. 6. The device of claim 1, wherein the sensor is capable of sending external radio signals capable of being captured by life guards or emergency services when the unconsciousness of the user is detected. 7. The device of claim 1, wherein the CO2 cartridge can be replaced by any other means of gas generation, for example decompression of a solid by heat. \\

\midrule

\textbf{Llama-3.1-8B-FT (Mixed)} 1. A device for a life preserver, comprising a collar (1) that is placed around the user's neck, said collar (1) being inflatable by means of a cartridge (2) of compressed gas, said collar (1) comprising a solenoid valve (3) in connection with said cartridge (2) for the filling of said collar (1), said collar (1) comprising a discharge valve (4) for the emptying of said collar (1) after its use, said collar (1) comprising a sensor (5) associated with a microprocessor (6) programmed with an algorithm that constantly receives information on the user's body functions and is capable of determining when the user is unconscious, said device comprising emission/reception means between said valve (3) and said microprocessor (6), said sensor (5) being able to send a signal that alters the threshold of the predetermined value and said microprocessor (6) to send a signal to said solenoid valve (3) of said cartridge (2) of compressed gas for the automatic filling, in such a way that this device keeps the user's head out of the water at the time in which the life preserver is used and, therefore, breathing can be maintained if the user is unconscious. 2. The device according to claim 1, characterised in that it comprises a cord to activate said cartridge (2) of compressed gas manually in case of need. 3. The device according to claim 1 or 2, characterised in that it comprises a rechargeable battery that keeps the electronic elements in operation. 4. The device according to any of the previous claims, characterised in that said microprocessor (6), programmed with said predetermined algorithm, detects by means of said sensor (5) an altered signal of the information obtain and sends a signal for the opening of said solenoid valve (3) for the activation of said cartridge (2). 5. The device according to any of the previous claims, characterised in that said sensor (5) includes a pulse oximeter that detects a low level of oxygen saturation in the blood, a cardiac sensor that detects an altered heart rate, a sensor that captures muscle contractions, a movement sensor, in addition to a depth sensor for athletes that practice water sports, in particular, surfing. 6. The device according to any of the previous claims, characterised in that it comprises a sensor capable of sending external radio signals capable of being captured by life guards or emergency services when the unconsciousness of the user is detected. \\
\midrule

\textbf{GPT-4o} 1. An inflatable life preserver device comprising:    - a collar configured to be placed around a user's neck;    - an inflatable life preserver housed within the collar;    - a CO2 cartridge connected to the inflatable life preserver for inflation;    - a solenoid valve operatively connected to the CO2 cartridge to control the release of CO2;    - a sensor configured to monitor the user's body functions;    - a microprocessor programmed with an algorithm to receive data from the sensor and determine when the user is unconscious;    - wherein the microprocessor sends a signal to the solenoid valve to inflate the life preserver upon detecting unconsciousness, thereby maintaining the user's head in a safe position above water.  2. The device of claim 1, wherein the sensor includes:    - a pulse oximeter for detecting oxygen saturation levels in the blood;    - a cardiac sensor for monitoring heart rate;    - a sensor for detecting muscle contractions;    - a movement sensor;    - a depth sensor for aquatic sports applications.  3. The device of claim 1, further comprising:    - a manual activation cord for the CO2 cartridge to allow manual inflation of the life preserver.  4. The device of claim 1, further comprising:    - a rechargeable battery to power the electronic components of the device.  5. The device of claim 1, wherein the sensor is further configured to send external radio signals to alert lifeguards or emergency services upon detecting the user's unconsciousness.  6. The device of claim 1, wherein the CO2 cartridge is replaceable with an alternative gas generation means, such as the decompression of a solid by heat.  7. The device of claim 1, wherein the collar is designed to be used in various sports, including underwater fishing and surfing, and is suitable for children or individuals with limited mobility.  8. The device of claim 1, wherein the microprocessor is configured to adjust the threshold values for detecting unconsciousness based on the sensor data.  9. The device of claim 1, wherein the inflatable life preserver includes a discharge valve for deflating the collar after use.  10. The device of claim 1, wherein the microprocessor and sensor are configured to continuously monitor and analyze the user's physiological data to ensure timely activation of the life preserver. \\
\midrule

\textbf{Llama-3.1-8B-FT (Mixed-DPO)} 1. A device (1, 2, 3, 4, 5, 6) for a user, comprising a collar (1) configured to be placed around a neck of the user, the collar (1) comprising a compressed inflatable life preserver (1) and a sensor (5) associated with a microprocessor (6) configured to be programmed with an algorithm, wherein the inflatable life preserver (1) is configured to be inflated by a cartridge (2) (2) comprising compressed CO2, wherein the inflatable life preserver (1) comprises a solenoid valve (3) in connection with the cartridge (2) for filling the inflatable life preserver (1) with CO2 from the cartridge (2), wherein the inflatable life preserver (1) comprises a discharge valve (4) for emptying the inflatable life preserver (1) after its use, wherein the sensor (5) is configured to constantly receive information on the user's body functions, wherein the microprocessor (6) is configured to determine, based on the received information, when the user is unconscious, wherein the device comprises emission/reception means between the solenoid valve (3) and the microprocessor (6), wherein the sensor (5) is configured to send a signal to the microprocessor (6) and the microprocessor (6) is configured to send a signal to the solenoid valve (3) to open the solenoid valve (3) for the automatic filling of the inflatable life preserver (1) with CO2 from the cartridge (2) when the user is unconscious, wherein the sensor (5) comprises a pulse oximeter configured to detect a low level of oxygen saturation in the blood,  \\

\textit{Continue on next page} \\

\end{tabular}
\end{table*}

\begin{table*}[ht]
\centering
\footnotesize
\begin{tabular}{p{15.5cm}}
a cardiac sensor configured to detect an altered heart rate, a sensor configured to capture muscle contractions, a movement sensor and a depth sensor, wherein the sensor (5) is configured to send external radio signals capable of being captured by life guards or emergency services when the unconsciousness of the user is detected, and wherein the device comprises a rechargeable battery configured to keep the electronic elements in operation,characterized in that the sensor (5) is configured to send a signal that alters a threshold of a predetermined value of the algorithm, and in that the microprocessor (6) is configured to send a signal to the solenoid valve (3) of the cartridge (2) of CO2 for the automatic filling of the inflatable life preserver (1) with CO2 from the cartridge (2) when the user is unconscious, wherein the microprocessor (6), programmed with the predetermined algorithm, detects by means of the sensor (5) an altered signal of the information obtained and sends a signal for the opening of the solenoid valve (3) for the activation of the cartridge (2).
2. The device (1, 2, 3, 4, 5, 6) according to claim 1, wherein the sensor (5) is configured to send the external radio signals when the user is unconscious. 3. The device (1, 2, 3, 4, 5, 6) according to claim 1, wherein the sensor (5) is configured to send the external radio signals when the user is unconscious and the sensor (5) is configured to send the external radio signals when the user is in a real and imminent danger. 4. The device (1, 2, 3, 4, 5, 6) according to claim 1, wherein the sensor (5) is configured to send the external radio signals when the user is unconscious and the sensor (5) is configured to send the external radio signals when the user is in a real and imminent danger, wherein the sensor (5) is configured to send the external radio signals when the user is in a predetermined time under water. 5. The device (1, 2, 3, 4, 5, 6) according to claim 1, wherein the sensor (5) is configured to send the external radio signals when the user is unconscious and the sensor (5) is configured to send the external radio signals when the user is in a real and imminent danger, wherein the sensor (5) is configured to send the external radio signals when the user is in a predetermined time under water, wherein the sensor (5) is configured to send the external radio signals when the user is in a predetermined depth under water. \\

\bottomrule
\end{tabular}

\caption{Example model outputs of patent EP3676170B1.}
\label{table:example_ouput}
\end{table*}
\begin{table*}[!ht]
\centering
\footnotesize
\begin{tabular}{|p{0.95\linewidth}|}
\toprule
\textbf{Instructions:} \\
You will be given the draft claims and the referenced claims of the same patent. 
Your task is to rate the draft claims on eight metrics using the referenced claims as the gold standard, and each scored from 1 to 10. 
Please make sure you read and understand these instructions carefully. Please keep this document open while reviewing, and refer to it as needed. \\
\\
\textbf{Evaluation Criteria:} \\

1. Coverage: Does the candidate claim fully capture all essential technical features described in the referenced claim? \\
2. Technical Precision: Does the candidate claim provide precise definitions and clear characterizations of technical features, compared to the referenced claim? \\   
3. Specificity: Does the candidate claim explicitly describe technical features in concrete detail rather than using overly abstract or generalized language, compared to the referenced claim?  \\
4. Clarity: Is the candidate claim logically coherent, unambiguous, and clearly understandable, compared to the referenced claim?  \\
5. Dependency Structure: Does the candidate claim exhibit correct hierarchical dependency relationships, compared to the referenced claim?  \\
6. Legal Terminology: Does the candidate claim correctly and professionally use patent-standard legal terms and expressions, compared to the referenced claim?  \\
7. Scope of Protection: Does the candidate claim achieve an appropriate scope of protection, compared to the referenced claim?  \\
\\
\textbf{Evaluation Steps:} \\
1. Read the referenced claims carefully and assume the referenced claims have scores of 10 in all Evaluation Criteria. \\
2. Read the draft claims and compare it to the referenced claims. \\
3. Assign a score from 1 to 10 for each metric based on the Evaluation Criteria. Output the scores ONLY. \\
\\
\textbf{Example:} \\
Referenced Claims: \texttt{<<Claims>>} \\
Draft Claims: \texttt{<<Claims>>} \\
Evaluation Form (scores ONLY): \\
- Coverage: X, \\
- Technical Precision: X, \\
- Specificity: X, \\
- Clarity: X, \\
- Conciseness: X, \\
- Dependency Structure: X, \\
- Legal Terminology: X, \\
- Scope of Protection: X. \\
            
\bottomrule
\end{tabular}

\caption{LLM-as-a-judge prompt used for patent claim evaluation. }
\label{tab:evalprompt}

\end{table*}

\end{document}